%% file: main.tex
\documentclass[runningheads]{llncs}

\usepackage{epsfig}
\usepackage{graphicx}
\usepackage{comment}

\usepackage{color}
\usepackage{amsmath}
\usepackage{amssymb}
\usepackage{bm}
\usepackage[square,sort,comma,numbers]{natbib}
\usepackage{algorithm,algpseudocode}
\usepackage{caption}
\usepackage{graphbox}
\usepackage{color}
\usepackage[dvipsnames]{xcolor}
\usepackage{widetext}
\usepackage{microtype}

\usepackage{booktabs}       
\usepackage{multirow}
\usepackage{wrapfig}
\usepackage{subcaption}
\newcommand{\etal}{\textit{et al}.}
\newcommand{\ie}{\textit{i}.\textit{e}.}
\newcommand{\eg}{\textit{e}.\textit{g}.}
\usepackage{comment}

\definecolor{orange}{rgb}{0.99,0.29,0.07}

\newcommand\emi{\textcolor{black}}
\newcommand\Gianni{\textcolor{black}}

\newcommand{\ab}[1]{\textcolor{black}{#1}}
\newcommand{\abc}[1]{\textcolor{black}{#1}}

\newcommand{\vomega}{\boldsymbol{\mathbf{\omega}}}
\newcommand{\vmu}{\boldsymbol{\mathbf{\mu}}}

\newcommand{\vSigma}{\boldsymbol{\mathbf{\Sigma}}}

\newcommand{\vv}{\mathbf{v}}
\newcommand{\vx}{\mathbf{x}}

\newcommand{\vz}{\mathbf{z}}
\newcommand{\bx}{\bm{x}}
\newcommand{\by}{\bm{y}}

\newcommand{\noise}{\varepsilon}
\newcommand{\mcd}{MC Dropout}
\newcommand{\de}{Deep Ensembles}

\definecolor{mygray}{gray}{0.5}
\newcommand{\gray}[1]{\textcolor{mygray}{#1}}
\newcommand{\purple}[1]{\textcolor{DarkOrchid}{#1}}
\newcommand{\ubold}[1]{\fontseries{b}\selectfont#1}

\begin{document}
\pagestyle{headings}
\mainmatter
\def\ECCVSubNumber{2682}  

\title{TRADI: Tracking deep neural network weight distributions for uncertainty estimation}

\titlerunning{TRADI Tracking DNN weight distributions}
%
\author{Gianni Franchi\inst{1,2} \and
Andrei Bursuc\inst{3} \and
Emanuel Aldea\inst{2} \and
S\'{e}verine Dubuisson\inst{4}   \and
Isabelle Bloch\inst{5} 
}
%
\authorrunning{Franchi et al.}
%

\institute{ENSTA  Paris, Institut polytechnique de Paris \and
SATIE, Universit\'{e}  Paris-Sud, Universit\'{e} Paris-Saclay \and
valeo.ai \and
CNRS, LIS, Aix Marseille University \and
LTCI, T\'{e}l\'{e}com Paris, Institut polytechnique de Paris \thanks{This work was supported by ANR Project MOHICANS (ANR-15-CE39-0005). We would like to thank Saclay-IA cluster and CNRS Jean-Zay supercomputer.}
}
\maketitle

\begin{abstract}
During training, the weights of a Deep Neural Network (DNN) are optimized from a random initialization towards a nearly optimum value minimizing a loss function. Only this final state of the weights is typically kept for testing, while the wealth of information on the geometry of the weight space, accumulated over the descent towards the minimum is discarded. In this work we propose to make use of this knowledge and leverage it for computing the distributions of the weights of the DNN. This can be further used for estimating the epistemic uncertainty of the DNN by
\ab{aggregating predictions from an ensemble of networks sampled from these distributions.}
To this end we introduce a method for tracking the trajectory of the weights during optimization, that does neither require any change in the architecture, nor in the training procedure. We evaluate our method\ab{, TRADI,} on standard classification and regression benchmarks, and on out-of-distribution detection for classification and semantic segmentation. 
We achieve competitive results, while preserving computational efficiency in comparison to 
\ab{ ensemble approaches.}

\end{abstract}


\input{intro.tex}

\input{contrib.tex}

\input{related.tex}

\input{expes.tex}

\section{Conclusion}
In this work we propose a novel technique for computing the epistemic uncertainty of a DNN. TRADI is conceptually simple and easy to plug to the optimization of any DNN architecture. We show the effectiveness of TRADI over extensive studies and compare against the popular \mcd~ and the state of the art \de. Our method exhibits an excellent performance on evaluation metrics for uncertainty quantification, and in contrast to \de, for which the training time depends on the number of models, our algorithm does not add any significant cost over conventional training times.

Future works involve extending this strategy to new tasks, \eg, object detection, or new settings, \eg, active learning. Another line of future research concerns transfer learning. So far TRADI is starting from randomly initialized weights sampled from a given Normal distribution. In transfer learning, we start from a pre-trained network where weights are expected to follow  a different distribution. If we have access to the distribution of the DNN weights we can improve the effectiveness of transfer learning with TRADI. 

\clearpage
%
%
\bibliographystyle{splncs04}
\bibliography{references}

\clearpage
\begin{widetext}
\begin{center}
\textbf{\large TRADI: Tracking deep neural network weight distributions (Supplementary material)}
\end{center}
\end{widetext}


\setcounter{equation}{0}
\setcounter{figure}{0}
\setcounter{table}{0}
\renewcommand{\theequation}{S\arabic{equation}}
\renewcommand{\thefigure}{S\arabic{figure}}

\input{content_supplementary.tex}

\end{document}

%% file: intro.tex
\section{Introduction}

\ab{In recent years, Deep Neural Networks (DNNs) have gained prominence in various computer vision tasks and practical applications. This progress has been in part  accelerated by multiple innovations in key parts of DNN pipelines, \eg, architecture design~\cite{krizhevsky2012imagenet,simonyan2014very, szegedy2014going, he2016deep}, optimization~\cite{kingma2014adam}, initialization~\cite{glorot2010understanding,he2015delving}, regularization~\cite{srivastava2014dropout,ioffe2015batch}, \emph{etc}., along with a pool of effective heuristics identified 
by practitioners. 
Modern DNNs achieve now strong accuracy across tasks and domains, leading to their potential utilization as key blocks in real-world applications.}

\ab{However, DNNs have also been shown to be making mostly over-confident predictions~\cite{guo2017calibration}, a side-effect of the heuristics used in modern DNNs. This means that for ambiguous instances bordering two classes (\eg, human wearing a cat costume), or on unrelated instances (\eg, plastic bag not ``seen" during training and classified with high probability as rock), DNNs are likely to fail silently, which is a critical drawback for decision making systems. This has motivated several works to address the predictive uncertainty of DNNs~\cite{blundell2015weight, gal2016dropout, lakshminarayanan2017simple}, usually taking inspiration from Bayesian approaches.
Knowledge about the distribution of the network weights during training opens the way for studying the evolution of the underlying covariance matrix, and the uncertainty of the model parameters, referred to as the epistemic uncertainty~\cite{kendall2017uncertainties}. 
In this work we propose a method for estimating the distribution of the weights by tracking their trajectory during training. This enables us to sample an ensemble of networks and estimate more reliably the epistemic uncertainty and detect out-of-distribution samples.
}
\begin{figure}[t!]
  \renewcommand{\figurename}{Fig.}
  \renewcommand{\captionlabelfont}{\bf}
  \renewcommand{\captionfont}{\small}
  \centering
  \begin{minipage}{0.45\textwidth}
  \includegraphics[width=0.9\textwidth, clip=true, trim = 0mm 0mm 0mm 30mm]{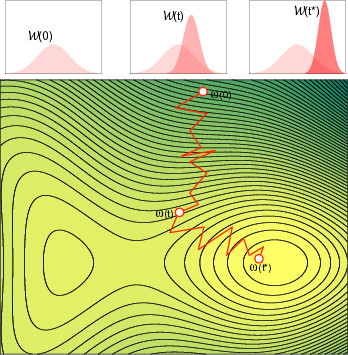}
  \end{minipage}
  \begin{minipage}{0.45\textwidth}
  \includegraphics[width=0.99\linewidth]{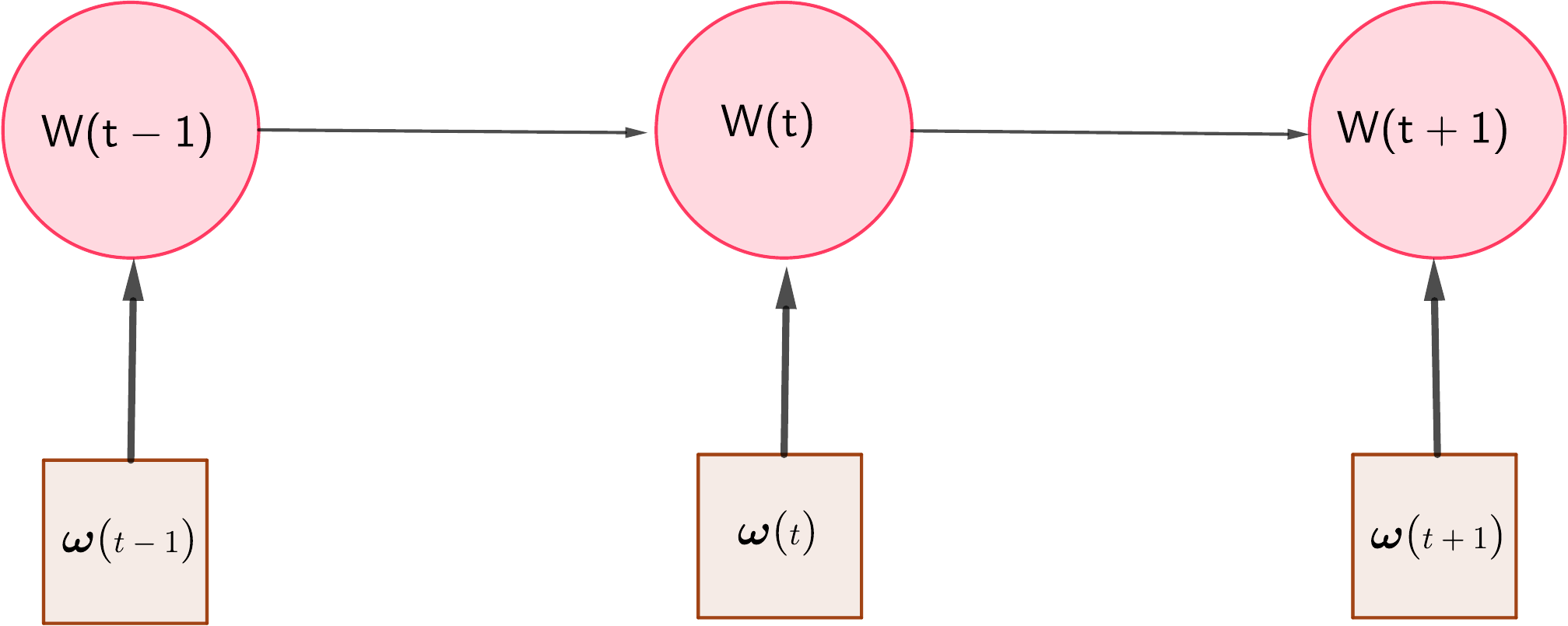}\par
  \vspace{2mm}
  \includegraphics[width=.99\textwidth, clip=true, trim = 0mm 99mm 0mm 0mm]{images/cvpr.jpg}
  \end{minipage}
   \caption{Our algorithm uses Kalman filtering for tracking the distribution $\mathcal{W}$ of all DNN weights across training steps from a generic prior $\mathcal{W}(0)$ to the final estimate $\mathcal{W}(t^*)$. We also estimate the covariance matrix of all the trainable network parameters. Popular alternative approaches rely typically either on ensembles of models trained independently \cite{lakshminarayanan2017simple} with a significant computational cost, approximate ensembles~\cite{gal2016dropout} or on averaging weights collected from different local minima~\cite{izmailov2018averaging}.
}
\label{fig:teaser}
\end{figure}

\ab{The common practice in training DNNs is to first initialize its weights using an appropriate random initialization strategy and then slowly adjust the weights through optimization according to the correctness of the network predictions on many mini-batches of training data. Once the stopping criterion is met, the final state of the weights is kept for evaluation. We argue that the trajectory of weights towards the (local) optimum reveals abundant information about the structure of the weight space that we could exploit, instead of discarding it and looking only at the final point values of the weights. Popular DNN weight initialization techniques~\cite{he2015delving,glorot2010understanding} consist of an effective layer-wise scaling of random weight values sampled from a Normal distribution. 
Assuming that weights follow a Gaussian distribution at time $t=0$, owing to the central limit theorem 
weights will also converge towards a Gaussian distribution. 
The final state is reached here through a noisy process, where the  stochasticity is induced by the weight initialization, the order and configuration of the mini-batches, \emph{etc}. We find it thus reasonable to see optimization as a random walk leading to a (local) minimum, in which case ``tracking'' the distribution makes sense (Fig.~\ref{fig:teaser}). To this end, Kalman filtering (KF)~\cite{grewal2011kalman} is an appropriate strategy for tractability reasons, as well as for the guaranteed optimality as long as the underlying assumptions are valid (linear dynamic system with Gaussian assumption in the predict and update steps).\footnote{Recent theoretical works~\cite{yann} show connections between optimization and KF, enforcing the validity of our approach.} To the best of our knowledge, our work is the first attempt to use such a technique to track the DNN weight distributions, and subsequently to estimate its epistemic uncertainty.
}

\noindent\textbf{Contributions.} The keypoints of our contribution are: \textbf{(a)} this is the first work which filters in a tractable manner the trajectory of the entire set of trainable parameters of a DNN during the training process; 
\ab{\textbf{(b)} we propose a tractable approximation for estimating the covariance matrix of the network parameters;} 
\ab{\textbf{(c)} we achieve competitive or state of the art results on most regression datasets, and on out-of-distribution experiments our method is better calibrated on three segmentation datasets (CamVid~\cite{brostow2008segmentation} , StreetHazards~\cite{hendrycks2016baseline}, and BDD Anomaly ~\cite{hendrycks2016baseline}); \textbf{(d)} our approach strikes an appealing trade-off in terms of performance and computational time (training + prediction). 
}


%% file: contrib.tex
\section{TRAcking of the weight DIstribution (TRADI)}

\ab{In this section, we detail our approach to first estimate the distribution of the weights of a DNN at each training step, and then generate an ensemble of networks by sampling from the computed distributions at training conclusion.}


\subsection{Notations and hypotheses}


\begin{itemize}

    \item $X$ and $Y$ are  two random variables, with $X\sim \mathcal{P}_X$ and $Y\sim \mathcal{P}_Y$. 
    \ab{Without loss of generality we consider the observed samples $\{\vx_i\}_{i=1}^{n}$ as vectors and the corresponding labels $\{y_i\}_{i=1}^{n}$ as scalars (class index for classification, real value for regression).}
    From this set of observations, we derive a training set of $n_l$ elements and a testing set of $n_{\tau}$ elements: $n=n_l+ {n_{\tau}}$.


 \item Training/Testing sets {are} denoted respectively {by} $\mathcal{D}_l=(\vx_i,y_i)_{i=1}^{n_l}$,   $\mathcal{D}_{\tau}=(\vx_i,y_i)_{i=1}^{n_{\tau}}$.  
 {Data in  $\mathcal{D}_l$ and   $\mathcal{D}_{\tau}$ are assumed to be i.i.d. distributed} according to {their respective unknown joint distribution} $\mathcal{P}_l$ and $\mathcal{P}_{\tau}$.


%
\item The DNN is defined by a vector containing the $K$ trainable weights $\vomega=\{\omega_k\}_{k=1}^{K}$. 
During training, $\vomega$ is iteratively updated for each mini-batch and we denote by $\vomega(t)$ the state of the DNN at  iteration $t$ of the optimization algorithm, realization of the random variable $W(t)$. 
 Let $g$ denote the architecture of the DNN associated with these weights and $g_{\vomega(t)}(x_i)$ its output at $t$. The initial set of weights $\vomega(0)=\{\omega_k(0)\}_{k=1}^{K}$ follows $\mathcal{N}(0,\sigma^2_k)$, where the values $\sigma_k^2$ are fixed as in~\cite{he2015delving}.


\item $\mathcal{L}(\vomega(t),y_i)$ is the loss function used to measure the dissimilarity between the output $g_{\vomega(t)}(\vx_i)$ of the DNN  and the expected output $y_i$. Different loss functions can be considered depending on the type of task.

\item Weights on different layers are assumed to be independent of each another at all times. \emi{This assumption is not necessary from a theoretical point of view, yet we need it to limit the complexity of the computation. Many works in the related literature rely on such assumptions~\cite{graves2011practical}, and some take the assumptions even further, \textit{e.g.}~\cite{blundell15bnn}, one of the most popular modern BNNs, supposes that all weights are independent (even from the same layer).} 
{Each weight $\omega_k(t)$, $k=1,\ldots,K$, follows a non-stationary Normal distribution (i.e. $W_{k}(t) \sim\mathcal{N}(\mu_k(t),\sigma_k^2(t))$) whose two parameters are tracked.}

\end{itemize}

\subsection{TRAcking of the DIstribution (TRADI) of weights of a DNN }
\subsubsection{Tracking the mean and variance of the weights}
DNN optimization typically starts from a set of randomly initialized weights $\vomega(0)$.
Then, at each training step $t$, 
several SGD updates are performed from 
 \ab{randomly chosen mini-batches} towards minimizing the loss. 
This makes the trajectory of the weights vary \ab{or oscillate}, but not necessarily in the good direction each time~\cite{lan2019lca}.
Since gradients are averaged over mini-batches, we can consider that weight trajectories are averaged over each mini-batch. After a certain number of epochs, the DNN converges, \ie~it reaches a local optimum with a specific configuration of weights that will then be used for testing. 
However, this general approach for training does not consider the evolution of the distribution of the weights, which may be estimated from the training trajectory and from the dynamics of the weights over time.
In our work, we argue that the history of the weight evolution up to their final state is an effective tool for estimating the epistemic uncertainty. 

More specifically, our goal is to estimate, for all weights $\omega_k(t)$ of the DNN and at each training step $t$, $\mu_k(t)$ and $\sigma_k^2(t)$, the parameters of their normal distribution.
Furthermore, for small networks we can also estimate the covariance $\mbox{cov}(W_{k}(t),W_{k'}(t))$ for any pair of weights $(\omega_k(t),\omega_k'(t))$ at $t$ in the DNN (see supplementary material for details). 
To this end, we leverage mini-batch SGD in order to optimize the loss between two weight realizations. 
The loss derivative with respect to a given weight ${\omega_{k}(t)}$ over a mini-batch $B(t)$ is given by:
\begin{equation}\label{notation_loss_grad}
\nabla\mathcal{L}_{\omega_{k}(t)} = \frac{1}{|B(t)|} \sum _{ (\vx_i,y_i) \in B(t)}\frac{\partial \mathcal{L}(\vomega(t-1),y_i)}{\partial \omega_{k}(t-1)}
\end{equation}
Weights $\omega_{k}(t)$ are then updated as follows:
  \begin{eqnarray}\label{linear_network_derivative0}
\omega_{k}(t)=\omega_{k}(t-1)-\eta \nabla\mathcal{L}_{\omega_{k}(t)}
 \end{eqnarray}
with $\eta$ the learning rate.

The weights of DNNs are randomly initialized at $t=0$  by sampling $W_{k}(0)\sim \mathcal{N}(\mu_k(0),\sigma^2_k(0))$,  where the parameters of the distribution are set empirically on a per-layer basis~\cite{he2015delving}.
By computing the expectation of $\omega_k(t)$ in Eq.~(\ref{linear_network_derivative0}), and using its linearity property, we get:
\begin{eqnarray}\label{transition_mean_v1}
\mu_k(t)=\mu_k(t-1) -\mathbb{E}\left[\eta \nabla\mathcal{L}_{\omega_{k}(t)}\right]
\end{eqnarray}
We can see that $\mu_k(t)$ depends on $\mu_k(t-1)$ and on another function at time $(t-1)$: this shows that the means of the weights follow a Markov process.
 
As in~\cite{ andrychowicz2016learning, yang2019scaling} we assume 
\ab{that during back-propagation and forward pass weights to be independent.}
We then get: 
   \begin{eqnarray}\label{transition_moment2}
\sigma_k^2(t)=\sigma_k^2(t-1)+\eta^2 \mathbb{E}\left[(\nabla\mathcal{L}_{\omega_{k}(t)} )^2\right]-\eta^2 \mathbb{E}^2\left[\nabla\mathcal{L}_{\omega_{k}(t)} \right]
 \end{eqnarray}

 This leads to the following state and measurement equations for 
$\mu_{k}(t)$: 
     \begin{eqnarray}\label{state_equ1}
 \left\{
    \begin{array}{l}
        \mu_{k}(t)=\mu_k(t-1)-\eta \nabla\mathcal{L}_{\omega_{k}(t)}   + \noise_{\mu} \\
          \omega_{k}(t)=\mu_{k}(t)+ \tilde{\noise}_{\mu} 
    \end{array}
\right.
   \end{eqnarray}
 with $\noise_\mu$ being the state noise, and $\tilde{\noise}_{\mu}$ being the observation noise, as realizations of $\mathcal{N}(0, \sigma^2_{\mu})$ and $\mathcal{N}(0, \tilde{\sigma}^2_{\mu})$ respectively. 
   The state and measurement equations for the variance $\sigma_{k}$ are given by:
\begin{eqnarray}\label{state_equ2}
\left\{
\begin{array}{l}
    \sigma_{k}^2(t)=\sigma_{k}^2(t-1)+ \left( \eta \nabla\mathcal{L}_{\omega_{k}(t)}  \right)^2+ \noise_{\sigma}\\
      z_k(t)=\sigma_{k}^2(t)-\mu_{k}(t)^2+\tilde{\noise}_{\sigma} \\
      \mbox{ with } z_k(t) =\omega_{k}(t)^2
\end{array}
\right.
   \end{eqnarray}
 \normalsize  
    with $\noise_\sigma$ being the state noise, and $\tilde{\noise}_{\sigma}$ being the observation noise, as realizations of $ \mathcal{N}(0, \sigma^2_{\sigma})$ and $\mathcal{N}(0, \tilde{\sigma}^2_{\sigma})$, respectively. We ignore the square empirical mean of the gradient on the equation as in practice its value is below the state noise.

   
 \subsubsection{Approximating the covariance}\label{sec:cov}  
   
Using the measurement and state transition in 
Eq.~(\ref{state_equ1}-\ref{state_equ2}), 
we can apply a Kalman filter to track the state of each trainable parameter. As the computational cost for tracking the covariance matrix is 
significant, we propose to track instead only the variance of the distribution. 
For that, we approximate the covariance by employing a model inspired from Gaussian Processes~\cite{williams2006gaussian}. 
We consider the Gaussian model due to its simplicity and good results.
Let 
$\vSigma(t)$ denote the covariance of $W(t)$, and let $\vv(t)=\begin{pmatrix}  \sigma_{0}(t),\sigma_{1}(t),\sigma_{2}(t), \ldots, \sigma_{K}(t)\end{pmatrix}$ be a vector of size $K$ composed of the standard deviations of all weights at time $t$. 
The covariance matrix is approximated by 
$\hat{\vSigma}(t) = (\vv(t) \vv(t)^{T})\odot \bm{\mathcal{K}}(t)$, 
where 
$\odot$ is the Hadamard product, and  
$\bm{\mathcal{K}}(t)$ is the kernel corresponding to the $K\times K$ Gram matrix of the weights of the DNN, 
with the coefficient $(k,k')$ given by 
$\bm{\mathcal{K}}(\omega_k(t),\omega_{k'}(t))=\exp \left(-{\frac {\|\omega_k(t)-\omega_{k'}(t)\|^{2}}{2\sigma_{\text{rbf}} ^{2}}}\right)$.
The computational cost for storing and processing the kernel $\bm{\mathcal{K}}(t)$ is however 
prohibitive in practice as its complexity is quadratic in terms of the number of weights
(\eg, $K\approx 10^9$ in recent DNNs).

Rahimi and Recht~\cite{Rahimi2007} alleviate this problem by approximating non-linear kernels, \eg~Gaussian RBF, in an unbiased way using random feature representations.
Then, for any translation-invariant positive definite kernel $  \bm{\mathcal{K}(t)}$, for all $(\omega_k(t),\omega_{k'}(t))$, $
 \bm{\mathcal{K}}(\omega_k(t),\omega_{k'}(t))$ 
depends only
on $\omega_k(t)-\omega_{k'}(t)$. We can then approximate the matrix by:
$$\bm{\mathcal{K}}(\omega_k(t),\omega_{k'}(t))  {\equiv} 
\mathbb{E}_{}\left[\cos(\Theta \omega_k(t) + \Phi )\cos(\Theta \omega_{k'}(t)+ \Phi)\right]$$ 
where 
$\Theta\sim\mathcal{N}(0,\sigma_{ \text{rbf}}^2)$ 
(this distribution is the Fourier transform of the kernel  distribution) and $\Phi\sim\mathcal{U}_{[0,2\pi]}$.
In detail, we approximate the high-dimensional feature space by projecting over the following $N$-dimensional feature vector:
\footnotesize
\begin{eqnarray}\label{eq:eqz}	
\vz(\omega_k(t)) {\equiv} \sqrt{\frac{2}{N}}\begin{bmatrix}\cos(\theta_1\omega_k(t) + \phi_1), \ldots , \cos(\theta_N \omega_k(t)+ \phi_N))\end{bmatrix}^\top
\end{eqnarray}
\normalsize
where the $\theta_1,\ldots,\theta_N$ are i.i.d. from $\mathcal{N}(0,\sigma_{\mbox{\tiny rbf}}^2)$  
and $\phi_1,\ldots,\phi_N$ are i.i.d. from $\mathcal{U}_{[0,2\pi]}$. In this new feature space we can approximate 
 kernel  $\bm{\mathcal{K}}(t)$  by $\hat{\bm{\mathcal{K}}}(t)$ defined by:
\footnotesize
\begin{eqnarray}
\hat{\bm{\mathcal{K}}}(\omega_{k}(t),\omega_{k'}(t)) =\vz(\omega_{k}(t))^\top \vz(\omega_{k'}(t))
\end{eqnarray}
\normalsize

Furthermore, it was proved in~\cite{Rahimi2007} that the probability of having an error of approximation greater than $\epsilon \in \mathbb{R}^+$ depends on $\exp(-N\epsilon^2)/\epsilon^2$.
To avoid the Hadamard product of matrices of size $K\times K$, we evaluate $\bm{r}(\omega_k(t)) =\sigma_{k}(t)\bm{z}(\omega_{k}(t)) $, and the value at index $(k,k')$ of the approximate covariance matrix $\hat{\vSigma}(t)$ is given by:
\begin{eqnarray}\label{eq:eqr}
\hat{\vSigma}(t)(k,k') =\bm{r}(\omega_k(t))^\top\bm{r}(\omega_k(t)).
\end{eqnarray}   

 \subsection{Training the DNNs}
 \label{sec:dual}

In our approach, for classification we use the cross-entropy loss to get the log-likelihood similarly to~\cite{lakshminarayanan2017simple}. 
For regression tasks, we train over two losses sequentially and modify $g_{\vomega(t)}(\bx_i)$  to have two output heads: the classical regression output $\mu_{\mbox{\tiny pred}}(\bx_i)$ and the predicted variance of the output $\sigma_{\mbox{\tiny pred}}^2$.
 This modification is inspired by 
\cite{lakshminarayanan2017simple}.
The first loss is the MSE 
$\mathcal{L}_1(\vomega(t),\by_i)=\|g_{\vomega(t)}(\bx_i)-\by_i \|^2_2$ 
as used in the traditional regression tasks. The second loss is the negative log-likelihood (NLL)~\cite{lakshminarayanan2017simple} which reads:
\begin{eqnarray}
\mathcal{L}_2(\vomega(t),y_i) = \frac{1}{2\sigma_{\mbox{\tiny pred}}(\vx_i)^2} \| \mu_{\mbox{\tiny pred}}(\vx_i) - y_i \|^2 + \frac{1}{2}\log \sigma_{\mbox{\tiny pred}}(\vx_i)^2 
\end{eqnarray}

We first train with loss 
$\mathcal{L}_1(\vomega(t),y_i)$ until reaching a satisfying $\vomega(t)$. 
In the second stage we add the variance prediction head and start fine-tuning from $\vomega(t)$ with loss $\mathcal{L}_2(\vomega(t),y_i)$. 
\ab{In our experiments we observed that this sequential training is more stable as it allows the network to first learn features for the target task and then to predict its own variance, rather than doing both in the same time (which is particularly unstable in the first steps).}



%
   
 \subsection{TRADI training algorithm overview}

\Gianni{We detail the TRADI steps during training in Appendix, Section 1.3.}
For tracking purposes we must store $\mu_k(t)$ and $\sigma_k(t)$ for all the weights of the network. Hence, the method computationally lighter than Deep Ensembles, which has a training complexity scaling with the number 
\ab{of networks composing the ensemble}. In addition, TRADI 
can be applied to any DNN without any modification of the architecture, in contrast to \mcd~that requires adding dropout layers to the underlying DNN.
For 
clarity 
we define $ \mathcal{L}(\vomega(t),B(t)) = \frac{1}{|B(t)|} \sum _{ (x_i,y_i) \in B(t)}\ \mathcal{L}(\vomega(t),y_i)$. 
Here $\mathbf{P}_\mu$, $\mathbf{P}_\sigma$ are the noise covariance matrices of the mean and variance respectively and $\mathbf{Q}_\mu$, $\mathbf{Q}_\sigma$ are the optimal gain matrices of the mean and variance respectively. These matrices are used during Kalman filtering \cite{kalman1960new}. 

\subsection{TRADI uncertainty during testing}
After having trained a DNN, we can evaluate its uncertainty by sampling new realizations of the weights from to the tracked distribution. We call  $\tilde{\vomega}(t)=\{\tilde{\omega}_k(t)\}_{k=1}^{K}$ the vector of size $K$ containing these realizations. 
Note that this vector is different from $\vomega(t)$ since it is sampled from the distribution computed 
with TRADI, that does not correspond exactly to the DNN weight distribution.
 In addition, we note $\vmu(t)$ the vector of size $K$ containing the mean of all weights at time $t$. 

Then, two cases can occur. In the first case, we have access to the covariance matrix of the weights (by tracking or by an alternative approach) that we denote $\vSigma(t)$, and we simply sample new realizations of $W(t)$ using the following formula:  
\begin{eqnarray}
    \tilde{\vomega}(t)= \vmu(t) +   \vSigma^{1/2}(t) \times \bm{m}_1  
\end{eqnarray}
in which $\bm{m}_1$ is drawn from the multivariate Gaussian $\mathcal{N}(\bm{0}_K,\bm{I}_K)$, where $\bm{0}_K,\bm{I}_K$ are respectively the $K$-size zero vector and the $K\times K$ size identity  matrix.

When we deal with a 
DNN (the considered case in this paper), we are constrained for tractability reasons to approximate the covariance matrix following the random projection trick proposed in the previous section, and we  generate new realizations of $W(t)$ as follows:
\begin{eqnarray}
\tilde{\vomega}(t)= \vmu(t)  +   \bm{R}(\vomega(t))  \times \bm{m}_2 
\end{eqnarray} 
where $ \bm{R}(\vomega(t))$ is a matrix of size $K\times N$ whose  rows $k \in [1,K]$ contain the $\bm{r}(\omega_k(t))^\top $ defined in Section~\ref{sec:cov}. $\bm{R}(\vomega(t))$ depends on  $(\theta_1,\ldots,\theta_N)$  
and on $(\phi_1,\ldots,\phi_N)$ defined in Eq.(\ref{eq:eqz}).
$\bm{m}_2$ is drawn from the multivariate Gaussian $\mathcal{N}(\bm{0}_N,\bm{I}_N)$, where $\bm{0}_N,\bm{I}_N$ are respectively the zero vector of size $N$ and the identity matrix of size $N\times N$. 
Note that since $N \ll K$, computations are significantly accelerated.

 Then similarly to works in \cite{maddox2019simple,kendall2017uncertainties}, given input data $(\vx^*,y^*)\in \mathcal{D}_{\tau}$ from the testing set, we estimate the marginal likelihood as  Monte Carlo integration. First, a sequence $\{\tilde{\vomega}^j(t)\}_{j=1}^{N{_{\mbox{\tiny model}}}}$ of $N{_{\mbox{\tiny model}}}$ realizations of $W(t)$ is drawn 
 (typically,  $N{_{\mbox{\tiny model}}}=20$). 
Then, the marginal likelihood  of $y^*$  over  $W(t)$ is approximated by:
 
 \begin{eqnarray}
\mathcal{P}(y^*|x^*) = \frac{1}{N{_{\mbox{\tiny model}}}} \sum_{j=1}^{N{_{\mbox{\tiny model}}}}  \mathcal{P}(y^*|\tilde{\vomega}^j(t),\vx^*)
\end{eqnarray} 
 
For regression, we use the strategy from~\cite{lakshminarayanan2017simple} to compute the log-likelihood of the regression 
and consider that the outputs of the DNN applied on $\vx^*$ are the parameters $\{\mu^j_{\mbox{\tiny pred}}(\vx^*),(\sigma^j_{\mbox{\tiny pred}}(\vx^*))^2\}_{j=1}^{N{_{\mbox{\tiny model}}}}$ of a Gaussian distribution (see Section~\ref{sec:dual}).  Hence, the final output is the result of a mixture of $N{_{\mbox{\tiny model}}}$ Gaussian distributions $\mathcal{N}(\mu^j_{\mbox{\tiny pred}}(\vx^*),(\sigma^j_{\mbox{\tiny pred}}(\vx^*))^2)$. 
During testing, if the DNN has BatchNorm layers, we first update BatchNorm statistics of each of the sampled $\tilde{\vomega}^j(t)$ models, where $j \in [1, N{_{\mbox{\tiny model}}}]$~\cite{izmailov2018averaging}.

%% file: related.tex
\section{Related work}

Uncertainty estimation is an important aspect for any machine learning model and it has been thoroughly studied across years in statistical learning areas. In the context of DNNs a renewed interest has surged in dealing with uncertainty,
In the following we briefly review methods related to our approach. 

\noindent\textbf{Bayesian methods.} Bayesian approaches deal with uncertainty by identifying a distribution of the parameters of the model. The posterior distribution is computed from a prior distribution assumed over the parameters and the likelihood of the model for the current data. The posterior distribution is iteratively updated across training samples. The predictive distribution is then computed through Bayesian model averaging by sampling models from the posterior distribution. This simple formalism is at the core of many machine learning models, including neural networks. Early approaches from Neal~\cite{neal1996} leveraged Markov chain Monte Carlo variants for inference on Bayesian Neural Networks. However for modern DNNs with millions of parameters, such methods are 
intractable for computing the posterior distribution, leaving the lead to gradient based methods.

\noindent\textbf{Modern Bayesian Neural Networks (BNNs).} Progress in variational inference~\cite{kingma13vae} has enabled a recent revival of BNNs. 
Blundell~\etal~\cite{blundell2015weight} learn distributions over neurons via a Gaussian mixture prior. While such models are
easy to reason along, 
they are limited to rather medium-sized networks.
Gal and Ghahramani~\cite{gal2016dropout} suggest that Dropout~\cite{srivastava2014dropout} can be used to mimic a BNN by sampling different subsets of neurons at each forward pass during test time and use them as ensembles. 
\mcd~is currently the most popular instance of BNNs due to its speed and simplicity, with multiple recent extensions~\cite{gal2017concrete, teye2018, lambert2018}. 
However, the benefits of Dropout are more limited for convolutional layers, where specific architectural design choices must be made~\cite{kendall2015bayesian, mukhoti2018eval}.
A potential drawback of \mcd~concerns the fact that its uncertainty is not reducing with more training steps~\cite{osband2016, osband2018}.
TRADI is compatible with both fully-connected and convolutional layers, while uncertainty estimates are expected to improve with training as it relies on the Kalman filter formalism.


\noindent\textbf{Ensemble Methods.} Ensemble methods are arguably the top performers for measuring epistemic uncertainty, and are largely applied to various areas, \eg~active learning \cite{beluch2018power}.
Lakshminarayan~\etal~\cite{lakshminarayanan2017simple} propose training an ensemble of DNNs with different 
initialization seeds.
The major drawback of this method is its computational cost since one has to train multiple DNNs, 
a cost which is particularly high for computer vision architectures, \eg, semantic segmentation, object detection.
Alternatives to ensembles use a network with multiple prediction heads~\cite{lee2015m}, collect weight checkpoints from local minima and average them~\cite{izmailov2018averaging} or fit a distribution over them and sample networks~\cite{maddox2019simple}.
Although the latter approaches are faster to train than ensembles, their limitation is that the observations from these local minima are relatively sparse for such a high dimensional space and are less likely to capture the true distributions of the space around these weights. With TRADI we are mitigating these points as we collect 
weight statistics at each step of the
SGD optimization. Furthermore, our algorithm has a lighter computational cost than~\cite{lakshminarayanan2017simple} during training.

\noindent\textbf{Kalman filtering (KF).} The KF \cite{kalman1960new} is a recursive estimator  that constructs an inference of unknown variables given measurements over time.
With the advent of 
DNNs, researchers have tried integrating ideas from KF in DNN training: 
for SLAM using RNNs~\cite{ionet2018, haarnoja2016backprop}, optimization~\cite{wang2018batch}, DNN fusion~\cite{liu2019neural}. 
In our 
approach, we employ KF for keeping track of the statistics of the network during training such that at ``convergence'' we have a better coverage of the distribution around each parameter of a multi-million parameter DNN. The KF provides a clean and relatively easy to deploy formalism to this effect.

\noindent\textbf{Weight initialization and optimization.}
Most DNN initialization techniques \cite{glorot2010understanding,he2015delving} start from weights sampled from a Normal distribution, and further scale them according to the number of units and the activation function. BatchNorm~\cite{ioffe2015batch} stabilizes training by enforcing a Normal distribution of intermediate activations at each layer. WeightNorm~\cite{salimans2016weight} has a similar effect over the weights, making sure they are sticking to the initial distributions. From a Bayesian perspective the $L_2$ regularization, known as weight decay, is equivalent to putting a Gaussian prior over the weights~\cite{bishop2006pattern}. We also consider a Gaussian prior over the weights, similar to previous works~\cite{blundell2015weight,izmailov2018averaging} for its numerous 
properties, ease of use and natural compatibility with KF. Note that we use it only in the filtering in order to reduce any major drift in the estimation of distributions of the weights across training, while mitigating potential instabilities in SGD steps.

%% file: expes.tex
 \section{Experiments}   

\ab{We evaluate TRADI on a range of tasks and datasets. For regression 
, in line with prior works~\cite{lakshminarayanan2017simple, gal2016dropout}, 
we consider a toy dataset and the regression benchmark~\cite{hernandez2015probabilistic}. For classification we evaluate on MNIST~\cite{lecun1998gradient} and CIFAR-10~\cite{krizhevsky2009learning}. Finally, we address the Out-of-Distribution task for classification, on MNIST/notMNIST~\cite{lakshminarayanan2017simple}, and for semantic segmentation, on CamVid-OOD, StreetHazards~\cite{hendrycks2019anomalyseg}, and BDD-Anomaly~\cite{hendrycks2019anomalyseg}.
Unless otherwise specified, we use mini-batches of size 128 and Adam optimizer with fixed learning rate of 0.1 in all our experiments. 
}

\begin{figure}[t!]
\renewcommand{\captionfont}{\small}
\begin{center}
\begin{tabular}{c c c}
\includegraphics[width=0.22\columnwidth]{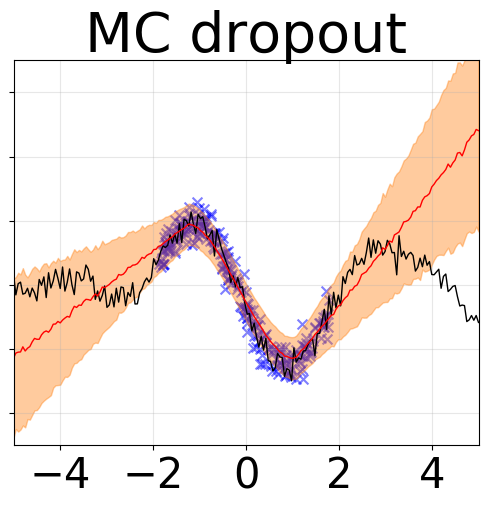}&
\includegraphics[width=0.22\columnwidth]{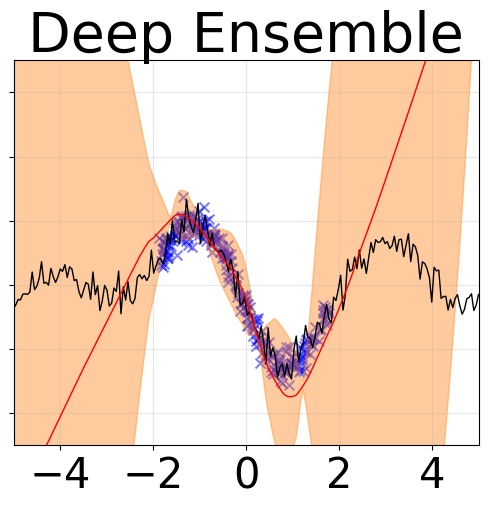}&
\includegraphics[width=0.22\columnwidth]{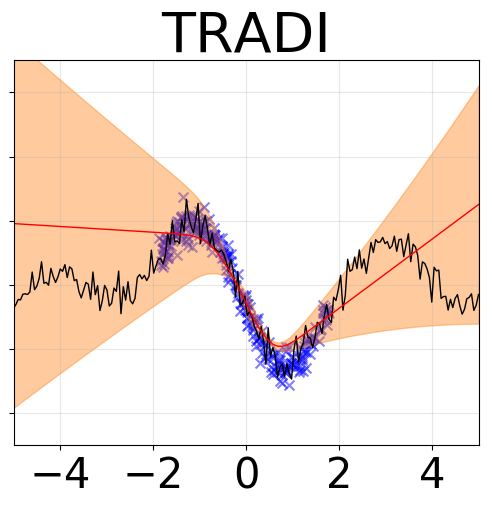}\\

\end{tabular}
\end{center}
\vspace{-0.5cm}
\caption{Results on a synthetic regression task comparing MC dropout, Deep Ensembles, and TRADI. $x$-axis: spatial coordinate of the Gaussian process. Black lines: ground truth curve. Blue points: training points. Orange areas: estimated variance. 
}\label{fig:regressiontoy}
\end{figure}

 \subsection{Toy experiments}

\ab{
\noindent \textbf{Experimental setup.}
As evaluation metric we use mainly the NLL uncertainty. In addition for classification we consider the  accuracy, while for regression we use the root mean squared error (RMSE).
For the out- of-distribution experiments we use the AUC, AUPR, FPR-95\%-TPR as in~\cite{hendrycks2016baseline}, and the Expected Calibration Error (ECE) as in~\cite{guo2017calibration}.
For our implementations we use PyTorch~\cite{paszke2019pytorch}. Unless otherwise specified, we use mini-batches of size 128 and Adam optimizer with fixed learning rate of 0.1 in all our experiments. We provide other implementation details on per-experiment basis.
}

First we perform a  qualitative evaluation 
on a one-dimensional synthetic dataset generated with a Gaussian Process of zero mean vector and as covariance function an RBF kernel $\bm{\mathcal{K}}$ with $\sigma^2=1$, denoted $GP(\bf{0},\bm{\mathcal{K}})$. We add to this process a zero mean Gaussian noise of variance  $0.3$. We train a neural network composed of one hidden layer and 200 neurons. In Fig.~\ref{fig:regressiontoy} we plot the regression estimation provided by TRADI, \mcd~\cite{gal2016dropout} and \de~\cite{lakshminarayanan2017simple}. Although $GP(\bf{0},\bm{\mathcal{K}})$ is one of the simplest stochastic processes, results show clearly that the compared approaches do not handle robustly the variance estimation, while TRADI neither overestimates nor underestimates the uncertainty.


 \subsection{Regression experiments}
 
For the regression task, we consider the experimental protocol and the data sets from \cite{hernandez2015probabilistic}, 
and also used in related works \cite{lakshminarayanan2017simple,gal2016dropout}. Here, we consider a neural network with one hidden layer, composed of 50 hidden units trained for 40 epochs. For each dataset, we do 20-fold cross-validation. For all datasets, we set the dropout rate to $0.1$ except for \textit{Yacht Hydrodynamics} and \textit{Boston Housing} for which it is set to $0.001$ and $0.005$, respectively. 
We compare against  \mcd~\cite{gal2016dropout} and \de~\cite{lakshminarayanan2017simple} and report  results in Table \ref{table:regression}. TRADI outperforms both methods, in terms of both RMSE and NLL. 
{Aside from the proposed approach to tracking the weight distribution, we assume that an additional reason for which our technique outperforms the alternative methods resides in the 
\ab{sequential} training (MSE and NLL) proposed in Section \ref{sec:dual}.}

\begin{table*}[t!]
\begin{center}
\renewcommand{\figurename}{Table}
\renewcommand{\captionlabelfont}{\bf}
\renewcommand{\captionfont}{\small} 
\caption{\small Comparative results on regression benchmarks}\label{table:regression}
\resizebox{0.80\columnwidth}{!}{
\begin{tabular}{   l  | c c c  | c c c}
\toprule
\multirow{2}{*}{Datasets} & \multicolumn{3}{c|}{RMSE} & \multicolumn{3}{c}{NLL} \\
& MC Dropout & Deep Ensembles & TRADI & MC Dropout & Deep Ensembles & TRADI \\ 
\midrule
Boston Housing      & $2.97 \pm 0.85$  & $3.28 \pm 1.00$  & $\bm{2.84  \pm  0.77}$  & $2.46 \pm 0.25$  & $2.41 \pm 0.25$  & $\bm{2.36 \pm 0.17}$ \\ 
\midrule
Concrete Strength   & $5.23 \pm 0.53$  & $6.03 \pm 0.58$  & $\bm{5.20 \pm 0.45}$  & $3.04 \pm 0.09$  & $3.06 \pm 0.18$  & $\bm{3.03 \pm 0.08}$ \\ 
\midrule
Energy Efficiency   & $1.66 \pm 0.16$  & $2.09 \pm 0.29$  & $\bm{1.20 \pm 0.27}$  & $1.99 \pm 0.09$  & $\bm{1.38 \pm 0.22}$  & $1.40 \pm 0.16$          \\ 
\midrule
Kin8nm              & $0.10 \pm 0.00$  & $0.09 \pm 0.00$  & $\bm{0.09 \pm 0.00}$  & $-0.95 \pm 0.03$ & $\bm{-1.2 \pm 0.02}$  & $-0.98 \pm 0.06$         \\ 
\midrule
Naval Propulsion    & $0.01 \pm 0.00$  & $0.00 \pm 0.00$  & $\bm{0.00 \pm 0.00}$  & $-3.80 \pm 0.05$ & $\bm{-5.63 \pm 0.05}$  & $-2.83 \pm 0.24$         \\ 
\midrule
Power Plant         & $4.02 \pm 0.18$  & $4.11 \pm 0.17$  & $\bm{4.02 \pm 0.14}$ & $2.80 \pm 0.05$  & $\bm{2.79 \pm 0.04}$  & $2.82 \pm 0.04$          \\ 
\midrule
Protein Structure   & $4.36 \pm 0.04$  & $4.71 \pm 0.06$  & $\bm{4.35 \pm 0.03}$ & $2.89 \pm 0.01$  & $2.83 \pm 0.02$  & $\bm{2.80 \pm 0.02}$ \\ 
\midrule
Wine Quality Red    & $0.62 \pm 0.04$  & $0.64 \pm 0.04$  & $\bm{0.62 \pm 0.03}$ & $0.93 \pm 0.06$  & $0.94 \pm 0.12$  & $\bm{0.93 \pm 0.05}$ \\ 
\midrule
Yacht Hydrodynamics & $1.11 \pm 0.38$  & $1.58 \pm 0.48$  & $\bm{1.05 \pm 0.25}$   & $1.55 \pm 0.12$  & $1.18 \pm 0.21$          & $\bm{1.18 \pm 0.39}$ \\ 
\bottomrule
\end{tabular}} 
\end{center}
\end{table*}

 \subsection{Classification experiments}

For the classification task, we conduct experiments on two datasets. The first one is the MNIST dataset \cite{lecun1998gradient}, which is composed of a training set containing 60k images and a testing set of 10k images, all of size $28\times28$. Here, we use a neural network with 3 hidden layers, each one containing 200 neurons, followed by ReLU non-linearities and BatchNorm, and fixed the learning rate $\eta = 10^{-2}$.  We share our results in Table \ref{table:classification}. 
For the MNIST dataset, we generate $N{_{\mbox{\tiny model}}} =20$ models, in order to ensure a fair comparison with \de.
The evaluation underlines that in terms of performance TRADI is positioned between \de~and \mcd. However, in contrast to \de~our algorithm is significantly lighter because only a single model needs to be trained, while \de~approximates the weight distribution by a very costly step of independent training procedures (in this case 20).

\begin{wraptable}{r}{7cm}
\vspace{-0.4cm}
\renewcommand{\figurename}{Table}
\renewcommand{\captionlabelfont}{\bf}
\renewcommand{\captionfont}{\small} 
\caption{\small Comparative results on image classification}\label{table:classification}
\vspace{-0.3cm}
\begin{center}
\resizebox{.4\columnwidth}{!}{
\begin{tabular}{   l  | c c  c c }
\toprule
\multirow{2}{*}{Method} & \multicolumn{2}{c}{ \ubold{MNIST}} & \multicolumn{2}{c}{\ubold{CIFAR-10}} \\
& NLL & ACCU & NLL & ACCU \\ 
\midrule
Deep Ensembles & \textbf{0.035} & \textbf{98.88} & 0.173 & 95.67 \\ 
\midrule
MC Dropout     & 0.065          & 98.19          & 0.205          & 95.27          \\
\midrule
SWAG         & 0.041          & 98.78          & \textbf{0.110}          & \textbf{96.41}         \\ 
\midrule
TRADI (ours)         & 0.044          & 98.63          & 0.205          & 95.29          \\ 
\bottomrule
\end{tabular}
}
\end{center}
\vspace{-0.5cm}
\end{wraptable} 

We conduct the second experiment on CIFAR-10~\cite{krizhevsky2009learning}, with WideResnet $28\times10$ ~\cite{zagoruyko2016wide} as DNN. The chosen optimization algorithm is SGD, $\eta=0.1$ and the dropout rate was fixed to $0.3$. Due to the long time necessary for \de~to train the DNNs we set $N{_{\mbox{\tiny model}}} =15$. Comparative results on this dataset, presented in Table \ref{table:classification}, allow us to make similar conclusions with experiments on the MNIST dataset.

 \subsection{Uncertainty evaluation for out-of-distribution (OOD) test samples.}
 
In \Gianni{these experiments, we evaluate uncertainty on OOD classes. We consider four datasets, and the objective of these experiments is to evaluate to what extent the trained DNNs are overconfident on instances belonging to classes which are not present in the training set. We report results in Table ~\ref{table:outofditribution}.} 

\ab{\noindent \textbf{Baselines.} We compare against \de~and \mcd, and propose two additional baselines. The first is the Maximum Classifier Prediction (MCP) which uses the maximum softmax value as prediction confidence and has shown competitive performance~\cite{hendrycks2016baseline, hendrycks2019anomalyseg}. Second, we propose a baseline to emphasize the ability of TRADI to capture the distribution of the weights. We take a \textit{trained} network and randomly perturb its weights with noise sampled from a Normal distribution. In this way we generate an ensemble of networks, each with different noise perturbations -- we practically sample networks from the vicinity of the local minimum. We refer to it as \textit{Gaussian perturbation ensemble}.  }

First we consider MNIST trained DNNs 
and use them on a test set composed of 10k MNIST images and 
\ab{19k images from NotMNIST~\cite{NotMnist}, a dataset of instances of ten classes of letters.}
Standard DNNs will assign letter instances of NotMNIST to a class number with high confidence as shown in~\cite{NotMnist}. For these OOD instances, our approach is able to decrease the confidence as illustrated in Fig.~\ref{fig:all}a, in which we represent the \textit{accuracy vs confidence} curves as in~\cite{lakshminarayanan2017simple}.

\begin{table}[t!]
\renewcommand{\figurename}{Table}
\renewcommand{\captionlabelfont}{\bf}
\renewcommand{\captionfont}{\small} 
\caption{Distinguishing in- and out-of-distribution data for semantic segmentation (CamVid, StreetHazards, BDD Anomaly) and image classification (MNIST/notMNIST)}
\label{table:outofditribution}
\begin{center}
\resizebox{0.88\columnwidth}{!}{
\begin{tabular}{ c l r r r r r }
\toprule
Dataset & OOD technique  & AUC  & AUPR & FPR-95\%-TPR & ECE & \ab{Train} time\\ 
\midrule
\multirow{5}{*}{\shortstack[c]{\ubold{MNIST/notMNIST} \\ 3 hidden layers}}      & Baseline (MCP) & 94.0 & 96.0 & 24.6    & \ubold{0.305}  & 2m   \\ 
   & Gauss. perturbation ensemble &  94.8 & 96.4 &  19.2  & 0.500  & 2m  \\ 
  &  MC Dropout      & 91.8 & 94.9 & 35.6   & 0.494  &   2m  \\ 
    & Deep Ensemble  & \ubold{97.2} & \ubold{98.0} & \ubold{9.2}     & 0.462  & 31m  \\ 
    & SWAG & 90.9 & 94.4 & 31.9     & 0.529  &  \\
   & TRADI (ours)   & 96.7 & 97.6 & 11.0    & 0.407  &    2m  \\ 
\midrule
\multirow{5}{*}{\shortstack[c]{\ubold{CamVid-OOD} \\ ENET}}                 & Baseline (MCP) & 75.4 & 10.0 & 65.1    &  0.146 & 30m  \\ 
    & Gauss. perturbation ensemble & 76.2 & 10.9  & 62.6    & 0.133  & 30m \\
   &  MC Dropout      & 75.4 & 10.7 & 63.2   &  0.168 & 30m \\ 
   & Deep Ensemble  & \ubold{79.7} & \ubold{13.0 }& \ubold{55.3}   & 0.112  & 5h \\
   & SWAG    & 75.6 & 12.1  &  65.8   & 0.133 &     \\
     & TRADI (ours)   & 79.3 & \ubold{12.8} & 57.7    & \ubold{0.110} & 41m    \\ 
\midrule
\multirow{5}{*}{\shortstack[c]{\ubold{StreetHazards} \\ PSPNet}}                 & Baseline (MCP) & 88.7 & 6.9  & 26.9    & 0.055 & 13h14m  \\ 
      & Gauss. perturbation ensemble &   57.08    & 2.4  &  71.0 & 0.185  & 13h14m  \\ 
    & MC Dropout      & 69.9 & 6.0  & 32.0   & 0.092  &   13h14m\\ 
    & Deep Ensemble  &\ubold{ 90.0} & 7.2  & 25.4    & 0.051  & 132h19m  \\ 
   & TRADI (ours) & 89.2 & \ubold{7.2}  & \ubold{25.3}  & \ubold{0.049 }&  15h36m     \\ \midrule
 \multirow{5}{*}{\shortstack[c]{\ubold{BDD Anomaly} \\ PSPNet}}   & Baseline (MCP) & 86.0 & 5.4  & 27.7    & 0.159  & 18h08 \\ 
      & Gauss. perturbation ensemble & 86.0 & 4.8  & 27.7    & 0.158  &  18h08m\\ 
      & MC Dropout      & 85.2 & 5.0  & 29.3    & 0.181  & 18h08m  \\
     & Deep Ensemble  & 87.0 & 6.0  & \ubold{25.0}    & 0.170  & 189h40m  \\ 
     &TRADI  (ours)   & 86.1 & 5.6  & 26.9     & \ubold{0.157}  & 21h48m   \\ 
\bottomrule
\end{tabular}
}
\end{center}
\end{table}

The \textit{accuracy vs confidence} curve is constructed by considering, for different confidence thresholds, all the test data for which the classifier reports a confidence above the threshold, and then by evaluating the accuracy on this data. The confidence of a DNN is defined as the maximum prediction score. \Gianni{We also evaluate the OOD uncertainty using AUC, AUPR and FPR-95\%-TPR metrics, introduced in \cite{hendrycks2016baseline} and the ECE metrics\footnote{Please note that the ECE is calculated over the joint dataset composed of the In distribution and the Out of distribution test data.} introduced in \cite{guo2017calibration}. These criteria characterize the quality of the prediction that a testing sample is OOD with respect to the training dataset. We also measured the computational training times of all algorithms implemented in PyTorch on a PC equipped with Intel Core i9-9820X and one GeForce RTX 2080 Ti
and report them in Table \ref{table:outofditribution}.} We note that TRADI DNN with 20 models provides incorrect predictions on such OOD samples with lower confidence than \de~and \mcd.

\begin{figure}[t!]
\renewcommand{\figurename}{Fig.}
\renewcommand{\captionlabelfont}{\bf}
\renewcommand{\captionfont}{\small}
     \centering
        \begin{subfigure}[b]{0.27\linewidth}
        \includegraphics[width=\textwidth]{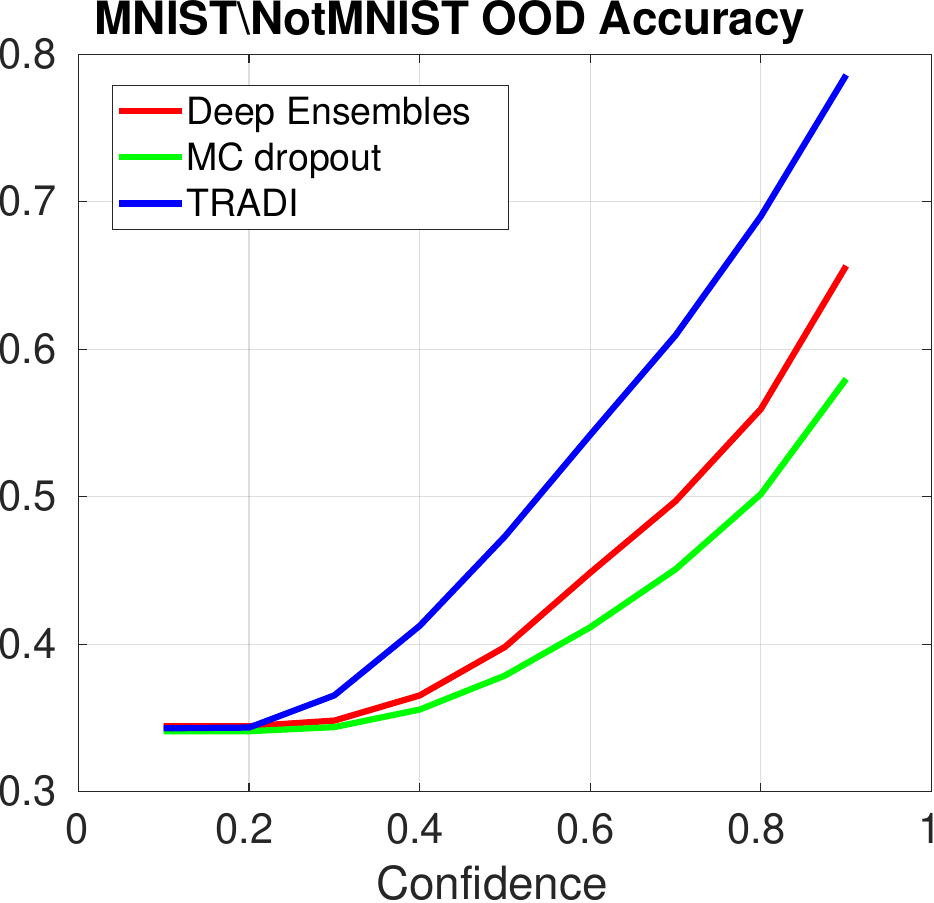}
        \label{mnistacc}
        \end{subfigure}\;
        \begin{subfigure}[b]{0.28\linewidth}
        \includegraphics[width=\textwidth]{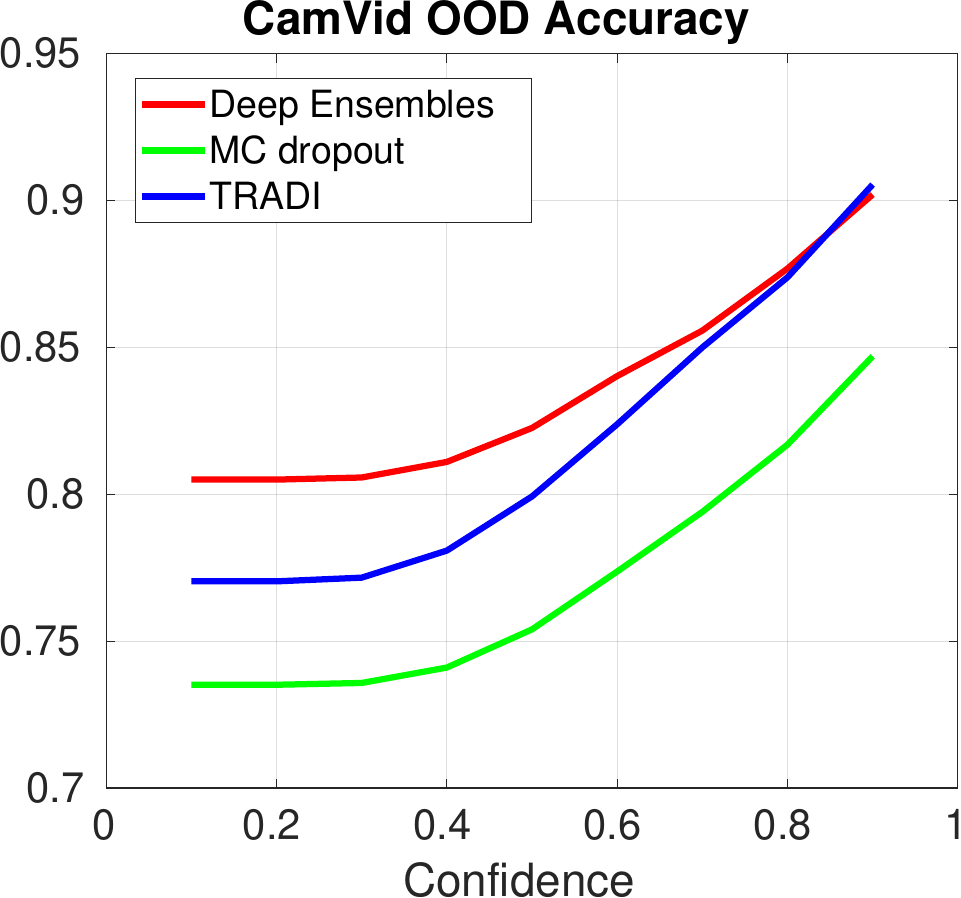}
        \label{camvidacc}
        \end{subfigure}\;
        \begin{subfigure}[b]{0.27\linewidth}
        \includegraphics[width=\textwidth]{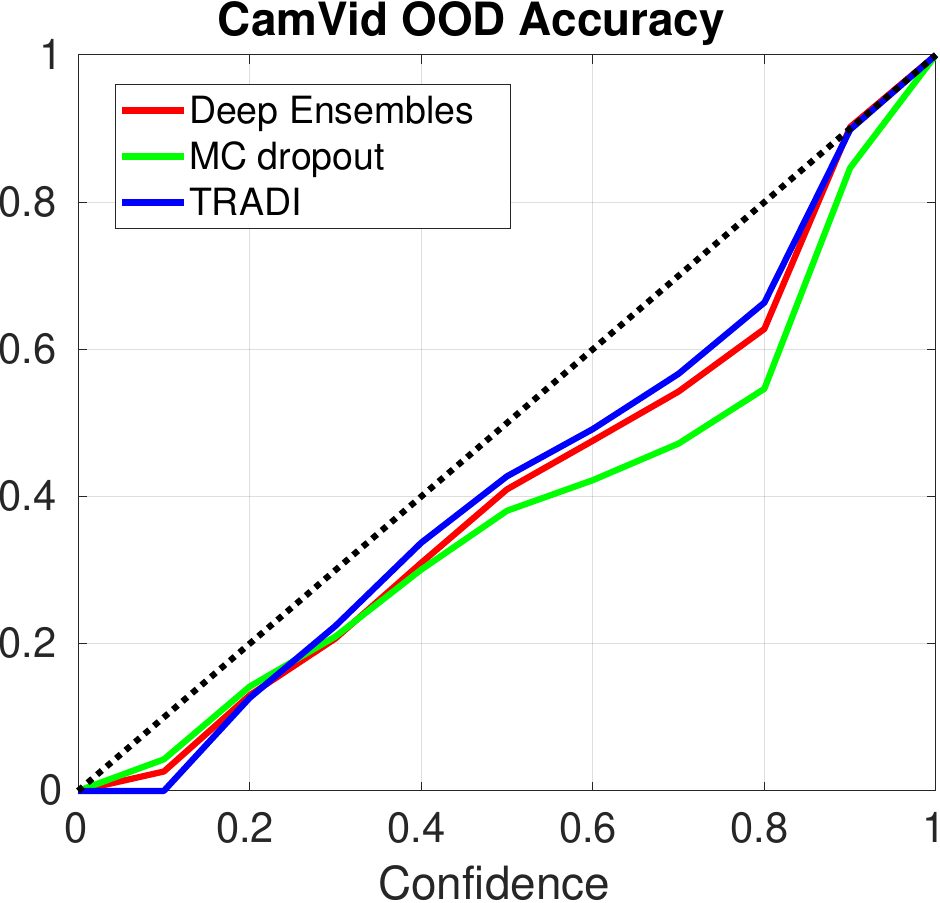}
        \label{camvidprec}
        \end{subfigure}\;
         \caption{(a) and (b) Accuracy vs confidence plot on the MNIST \textbackslash notMNIST and CamVid experiments, respectively.  (c) Calibration plot for the CamVid experiment.}
\label{fig:all}
\end{figure}

In the second experiment, we train a Enet DNN
~\cite{paszke2016enet} for semantic segmentation on CamVid dataset~\cite{brostow2008segmentation}. During training, we delete three classes (pedestrian, bicycle, and car), by marking the corresponding pixels as unlabeled. Subsequently, we test with data containing the classes represented during training, as well as the deleted ones. The goal of this experiment is to evaluate the DNN behavior on the deleted classes which represent thus OOD classes. \ab{We refer to this setup as CamVid-OOD.} In this experiment we use $N{_{\mbox{\tiny model}}}=10$ models 
trained for 90 epochs with SGD and using a learning rate $\eta=5\times10^{-4}$. In Fig. \ref{fig:all}b and \ref{fig:all}c we illustrate the \textit{accuracy vs confidence} curves and the \textit{calibration} curves \cite{guo2017calibration} for the CamVid experiment. The calibration curve as explained in \cite{guo2017calibration} consists in dividing the test set into bins of equal size according to the confidence, and in computing the accuracy over each bin. 
Both the calibration and the \textit{accuracy vs confidence} curves highlight whether the DNN predictions are good for different levels of confidence. However, the calibration provides a better understanding of what happens for different scores. 


\ab{Finally, we conducted experiments on the recent OOD benchmarks for semantic segmentation StreetHazards~\cite{hendrycks2019anomalyseg} and BDD Anomaly~\cite{hendrycks2019anomalyseg}. The former consists of 5,125/1,031/1,500 (train/test-in-distribution/test-OOD) synthetic images~\cite{dosovitskiy17} with annotations for 12 classes for training and a 13\textsuperscript{th} OOD class found only in the test-OOD set. The latter is a subset of BDD~\cite{yu2018bdd100k} and is composed of 6,688/951/361 images, with the classes \textit{motorcycle} and \textit{train} as anomalous objects. 
We follow the experimental setup from~\cite{hendrycks2019anomalyseg}, \ie, PSPNet~\cite{zhao2017pyramid} with ResNet50~\cite{he2016deep} backbone. On StreetHazards, TRADI outperforms \de~and on BDD Anomaly \de~has best results close to the one of TRADI.}

Results show that TRADI outperforms the alternative methods in terms of calibration, and that it may provide more reliable confidence scores. Regarding \textit{accuracy vs confidence}, the most significant results for a high level of confidence, typically above 0.7, show how overconfident the network tends to behave; in this range, our results are 
similar to those of \de. \Gianni{Lastly, in all experiments TRADI obtains performances close to the best AUPR and AUC, while having a computational time \ab{/training time significantly} smaller than \de.}

\begin{figure}[t!]
\renewcommand{\figurename}{Fig.}
\renewcommand{\captionlabelfont}{\bf}
\renewcommand{\captionfont}{\small}
\centering
\scalebox{0.8}
{
\begin{subfigure}{0.24\textwidth}\centering
  \caption{\scriptsize  Input + GT}
  \includegraphics[width=.98\textwidth]{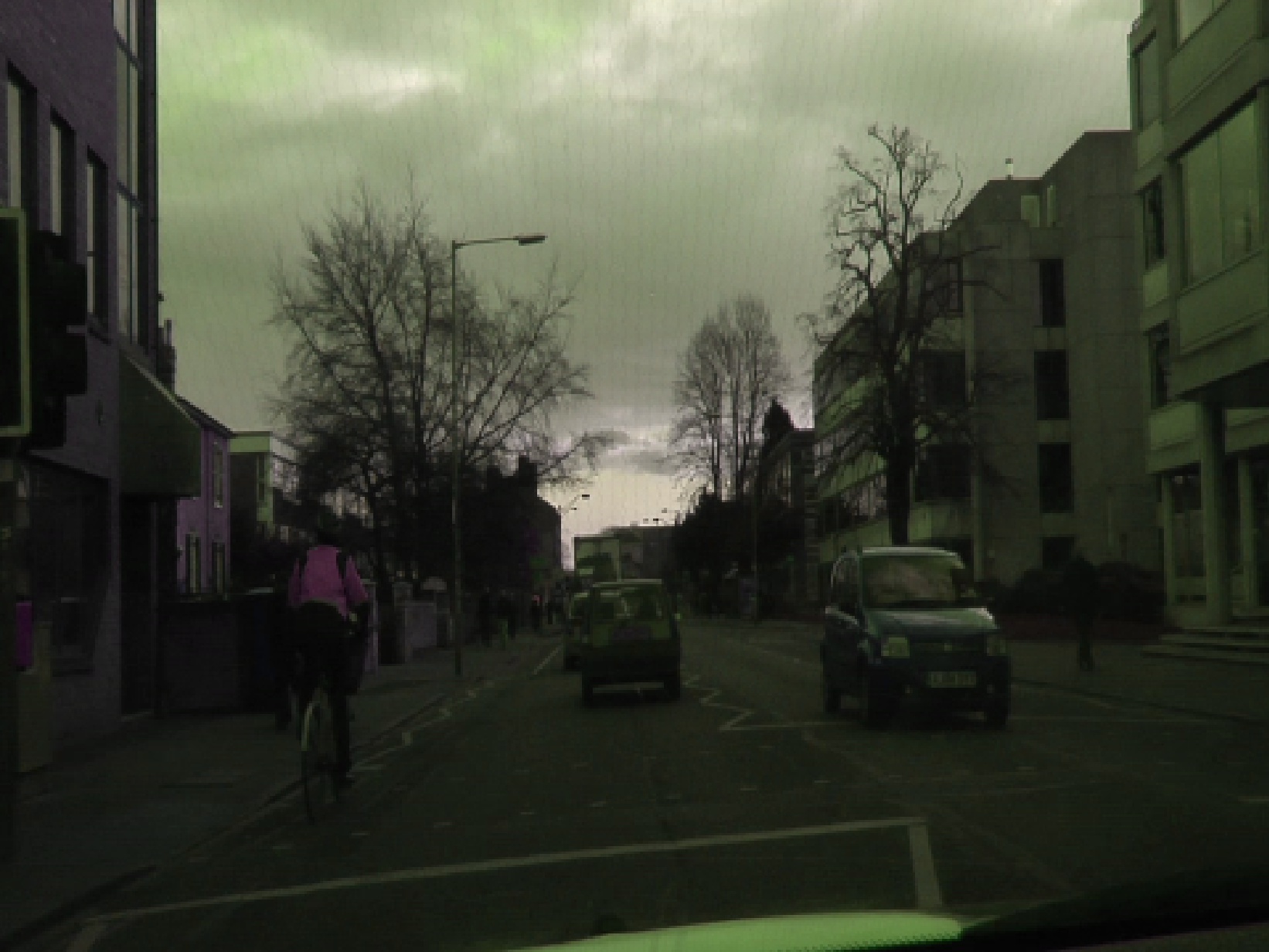}
\end{subfigure}
\begin{subfigure}{0.24\textwidth}
  \caption{\scriptsize MC Dropout}\centering
  \includegraphics[width=.98\textwidth]{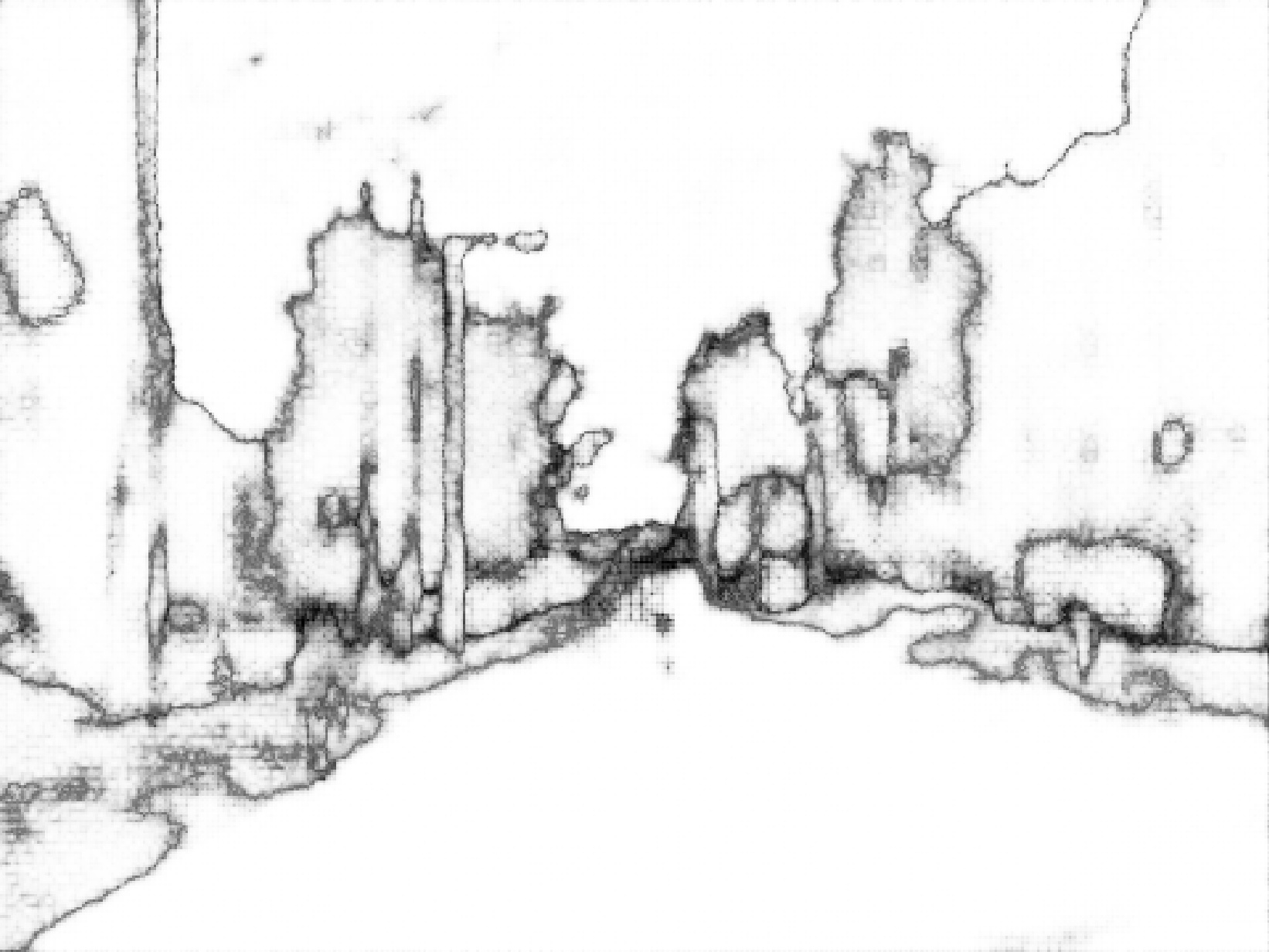}
\end{subfigure}
\begin{subfigure}{0.24\textwidth}\centering
  \caption{\scriptsize Deep Ensembles}
  \includegraphics[width=.98\textwidth]{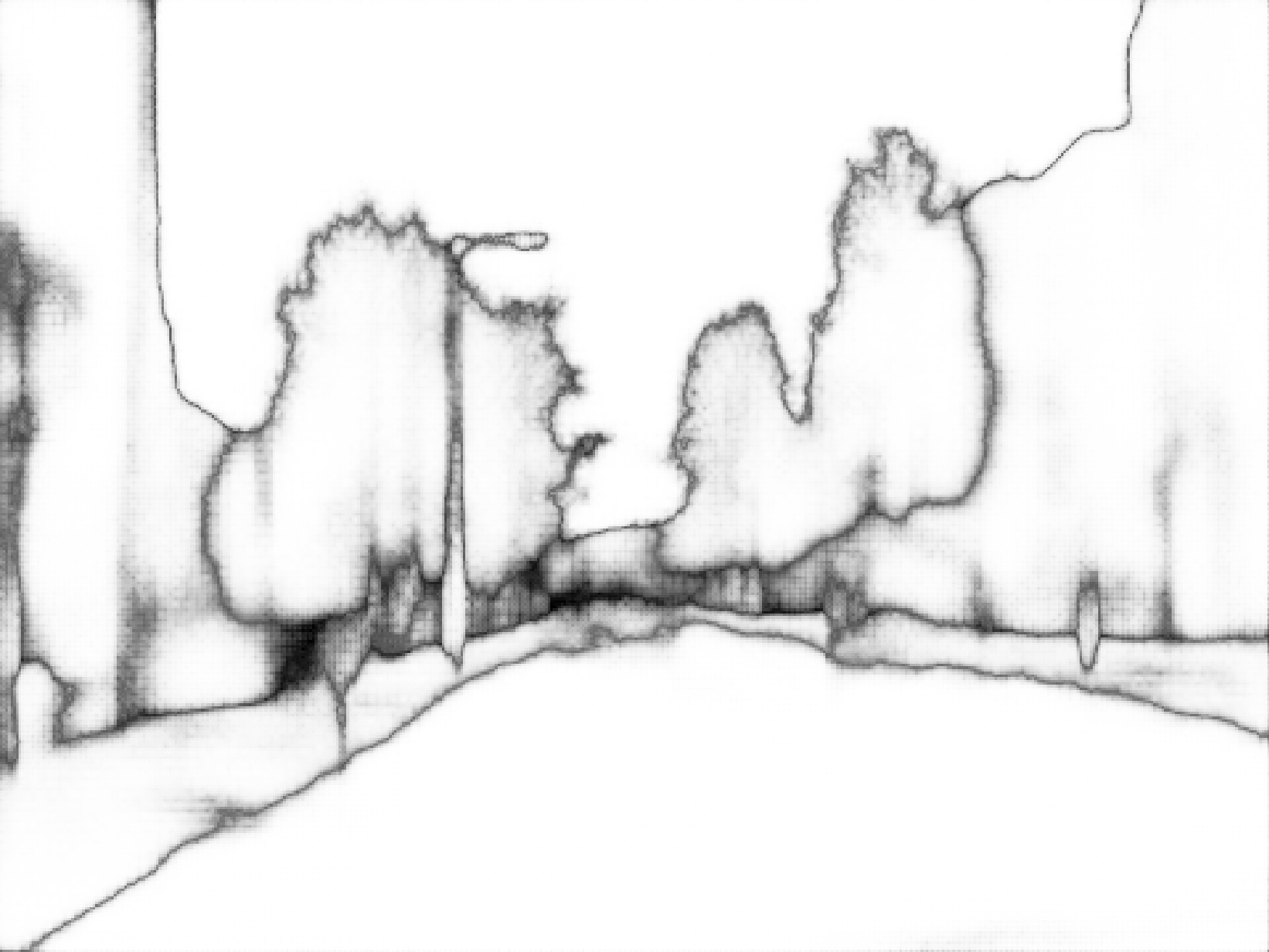}
\end{subfigure}
\begin{subfigure}{0.24\textwidth}\centering
  \caption{\scriptsize TRADI}
  \includegraphics[width=.98\textwidth]{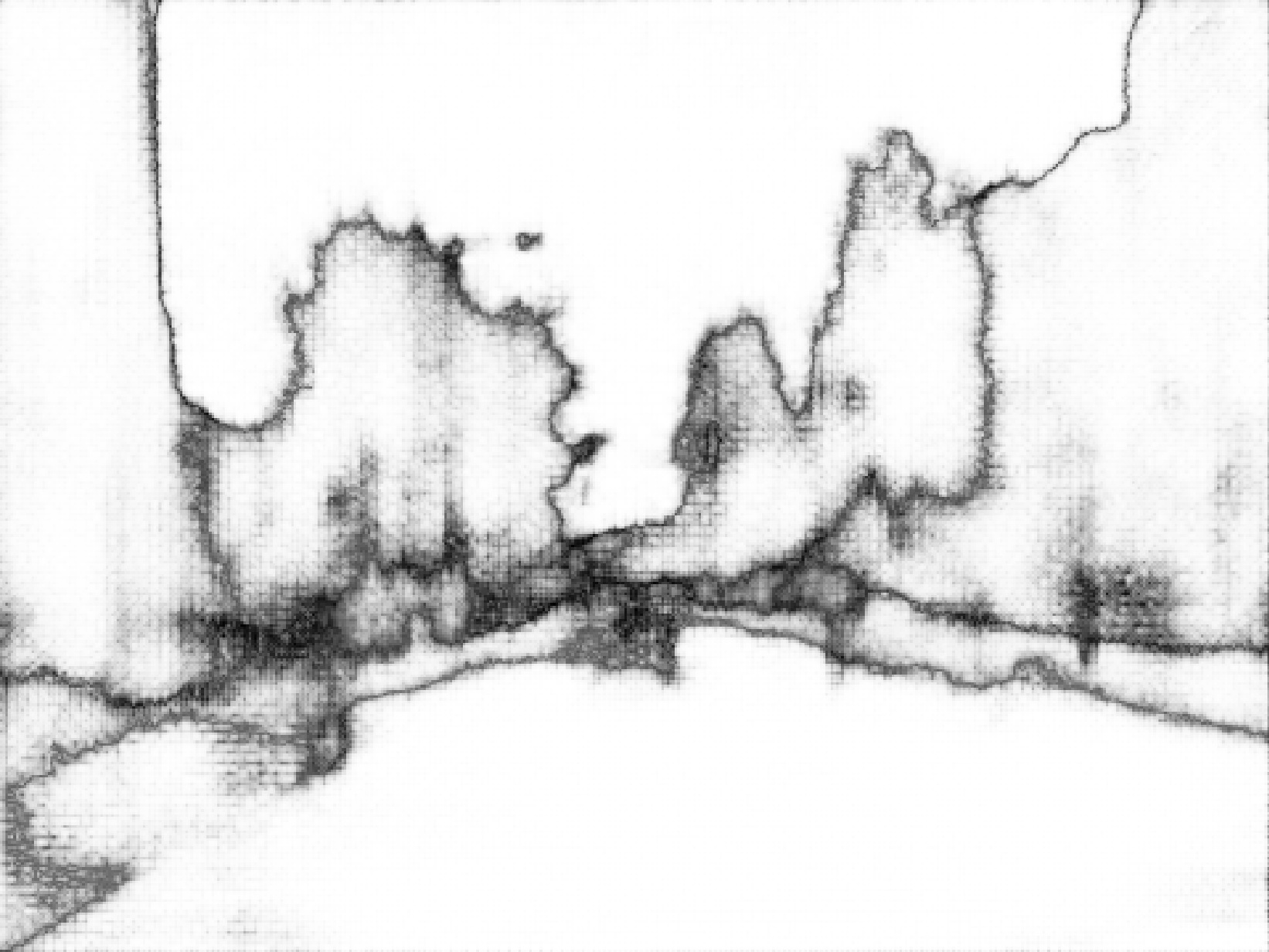}
\end{subfigure}
}
\scalebox{0.8}
{
\begin{subfigure}{0.24\textwidth}\centering
  \includegraphics[width=.98\textwidth]{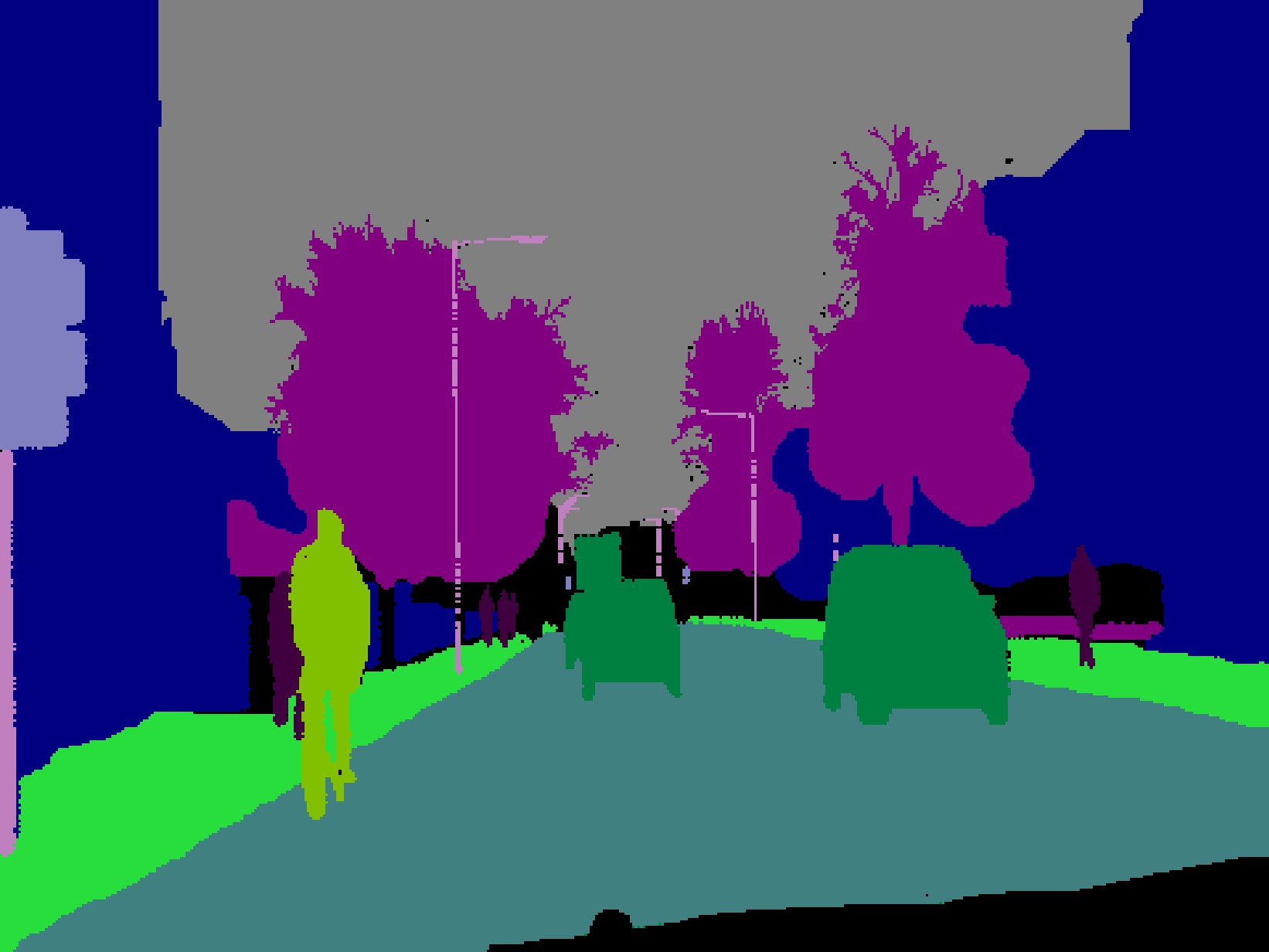}
\end{subfigure}
\begin{subfigure}{0.24\textwidth}\centering
  \includegraphics[width=.98\textwidth]{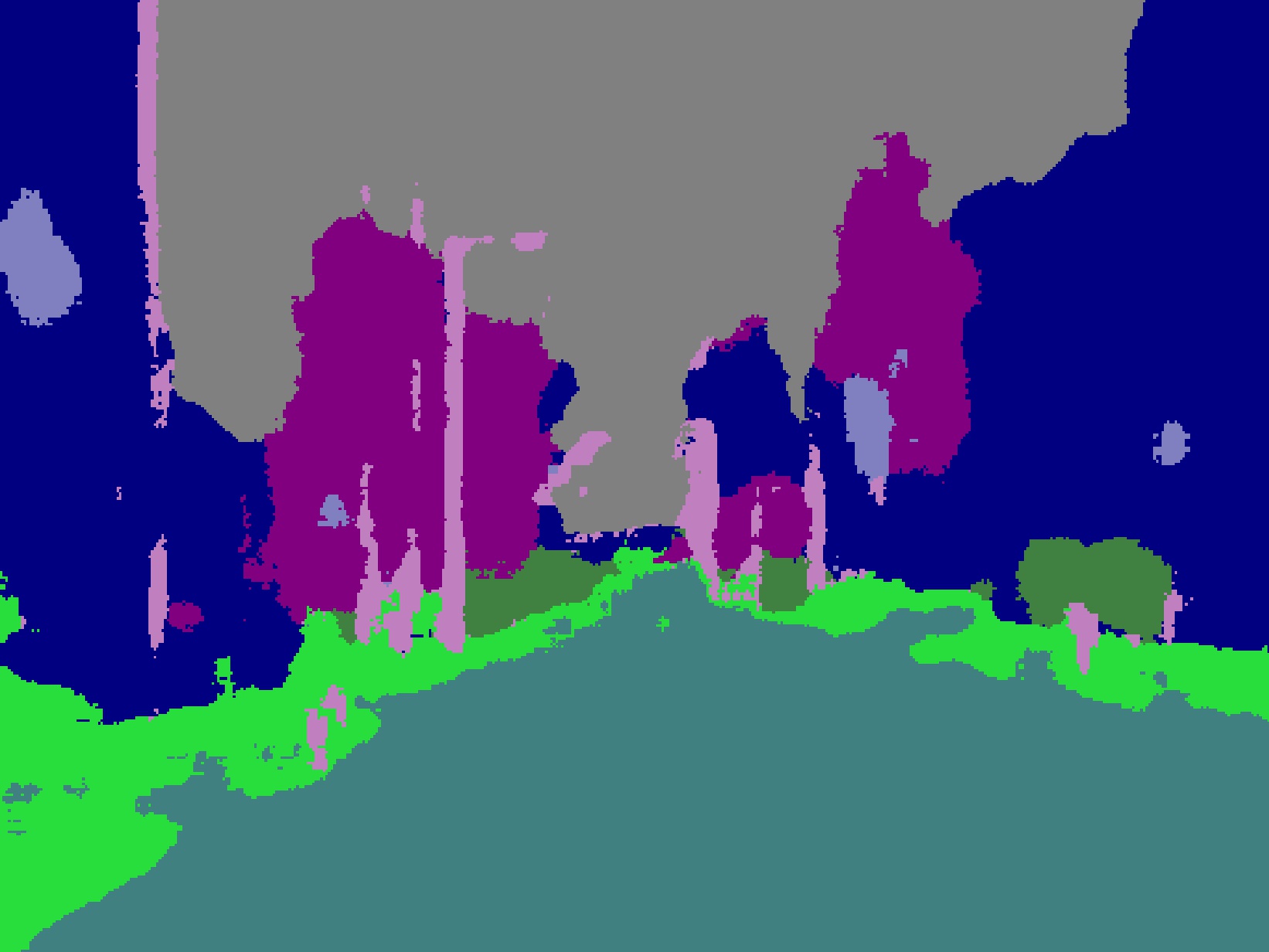}
\end{subfigure}
\begin{subfigure}{0.24\textwidth}\centering
  \includegraphics[width=.98\textwidth]{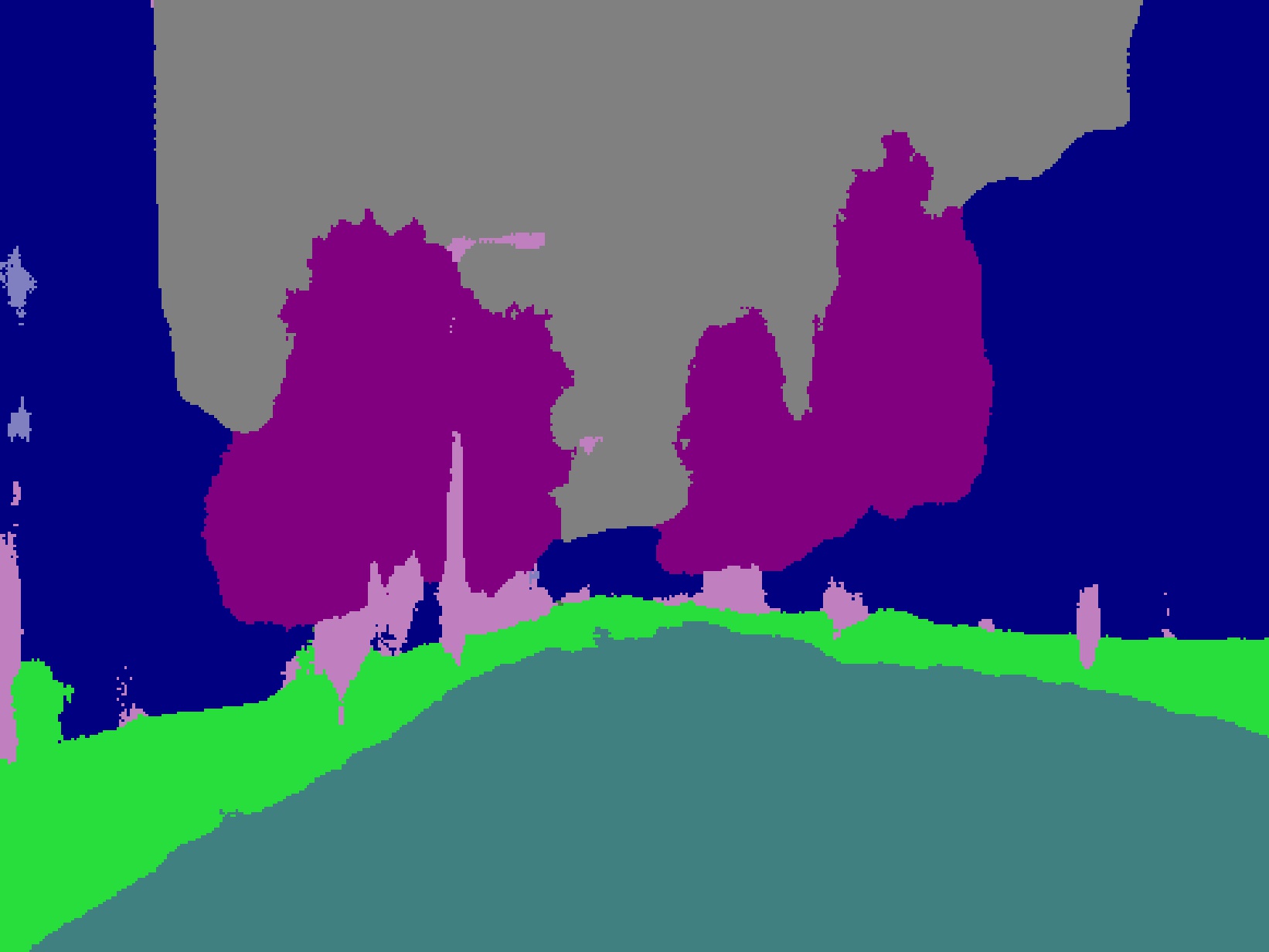}
\end{subfigure}
\begin{subfigure}{0.24\textwidth}\centering
  \includegraphics[width=.98\textwidth]{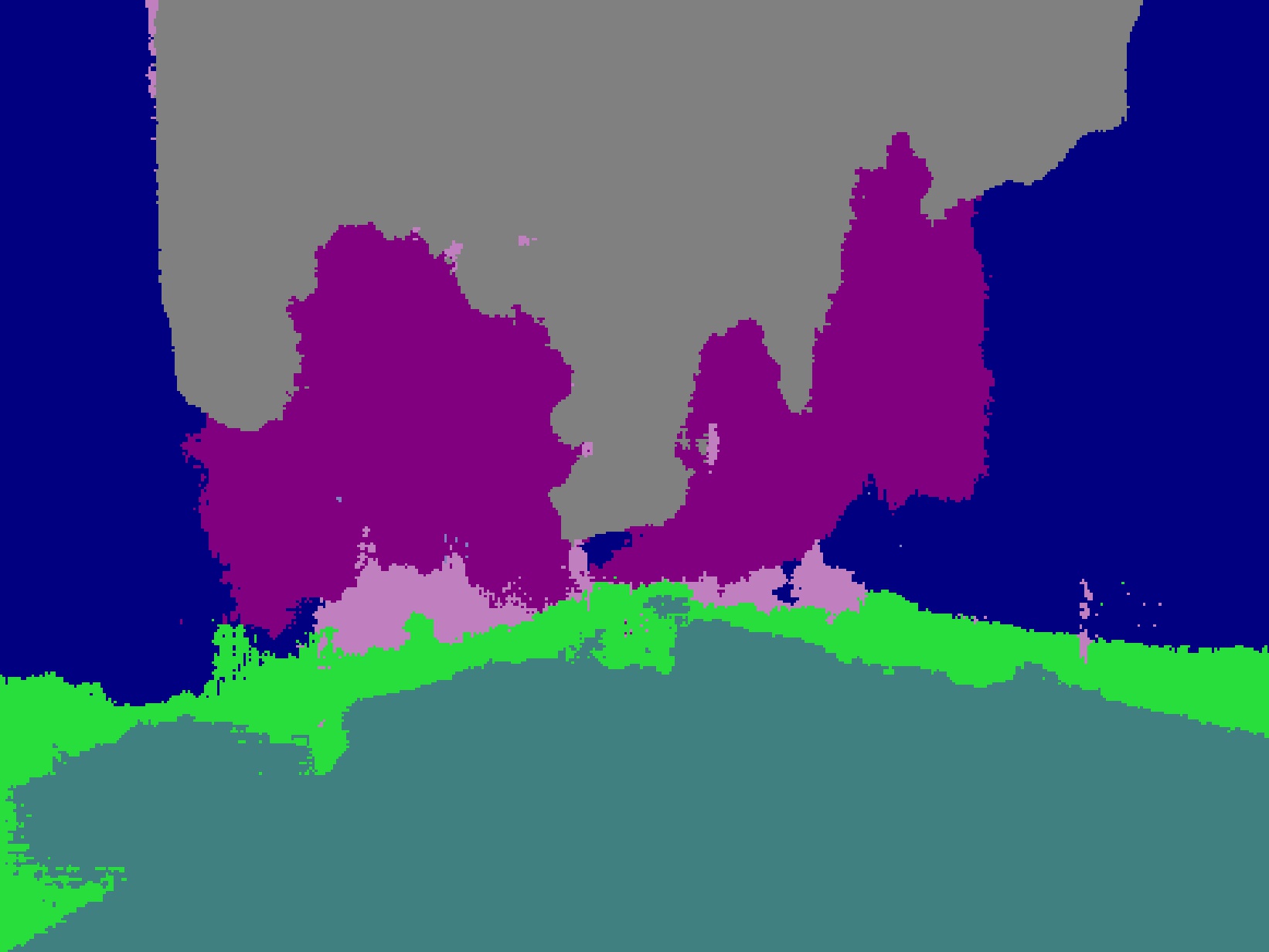}
\end{subfigure}
}
\scalebox{0.8}
{
\begin{subfigure}{0.24\textwidth}\centering
  \includegraphics[trim={17cm 4cm 0 7cm},clip,width=0.75\columnwidth]{./images/1_img.jpg}
\end{subfigure}
\begin{subfigure}{0.24\textwidth}\centering
  \includegraphics[trim={17cm 4cm 0 7cm},clip,width=0.75\columnwidth]{./images/1_confidence_drop.jpg}
\end{subfigure}
\begin{subfigure}{0.24\textwidth}\centering
  \includegraphics[trim={17cm 4cm 0 7cm},clip,width=0.75\columnwidth]{./images/1_confidence_deep_ense.jpg}
\end{subfigure}
\begin{subfigure}{0.24\textwidth}\centering
  \includegraphics[trim={17cm 4cm 0 7cm},clip,width=0.75\columnwidth]{./images/1_confidence_tradi.jpg}
\end{subfigure}
}
\caption{\small Qualitative results on CamVid\ab{-OOD}. Columns: (a) input image and ground truth; (b)-(d) predictions and confidence scores by MC Dropout, Deep Ensembles, and TRADI. Rows: (1) input and confidence maps; (2) class predictions; (3) zoomed-in area on input and confidence maps}
\label{fig:qualitative_results}
\end{figure}

\noindent \textbf{Qualitative discussion.} In Fig.~\ref{fig:qualitative_results} we 
give
as example a scene featuring the three OOD instances of interest (\textit{bike}, \textit{car}, \textit{pedestrian}). Overall, \mcd~outputs a noisy uncertainty map, but fails to highlight the OOD samples. By contrast,
\de~is overconfident, with  higher uncertainty values mostly around the borders of the objects. TRADI uncertainty is higher on borders and also on  pixels belonging to the actual OOD instances, as shown in the zoomed-in crop of the pedestrian in Fig.~\ref{fig:qualitative_results} (row 3). 

%% file: content_supplementary.tex
\section{TRAcking of the DIstribution (TRADI) of weights of a neural network }

This section details the tracking of the covariance matrix of a shallow neural network (see Section 3.2.1 of the main paper), gives the whole TRADI algorithm and explain how its parameters are chosen.

\subsection{Tracking the mean and variance of the weights and the covariance }

We consider a neural network (NN) for which each layer has few neurons, 
less than 100. Our goal here 
is to estimate, for all weights $\omega_k(t)$ of the NN and at each time step $t$ of the training process, $\mu_k(t)$ and $\sigma_k^2(t)$ the parameters of their normal distribution.  
Furthermore, we want to estimate  $\Sigma_{k,k'}(t)$ which is the covariance matrix between the $\omega_k(t)$
and  $\omega_{k'}(t)$. Note that, since we 
assume that weights on different layers are independent, we evaluate the covariance 
for $k,k'$ belonging to the same layer, otherwise their covariance is 
null.
To this end, we leverage mini-batch SGD to optimize the loss between two weight realizations. 

The derivative of the loss
with respect to a given weight  $\omega_k(t-1)$ 
over a mini-batch $B(t)$ is given by:
\begin{equation}\label{notation_loss_grad}
\nabla\mathcal{L}_{\omega_{k}(t)} = \frac{1}{|B(t)|} \sum _{ (x_i,y_i) \in B(t)}\frac{\partial \mathcal{L}(\vomega(t-1),y_i)}{\partial \omega_{k}(t-1)}
\end{equation}
Weights $\omega_{k}(t)$ are then updated as follows:
  \begin{eqnarray}\label{linear_network_derivative0}
\omega_{k}(t)=\omega_{k}(t-1)-\eta \nabla\mathcal{L}_{\omega_{k}(t)}
 \end{eqnarray}

The weights of NNs are randomly initialized at $t=0$  by sampling $W_{k}(0)\sim \mathcal{N}(\mu_k(0),\sigma^2_k(0))$,  where the parameters of the distribution are set empirically on a per-layer basis as in~\cite{he2015delving}. In addition, for all couples of weights $(k,k')$, the corresponding element of the covariance matrix is given by $\Sigma_{k,k'}(0)=0$  since all the weights are considered independent at time $t=0$.

Similarly with the main article, we use the following state and measurement equations for the mean $\mu_{k}(t)$: 
     \begin{eqnarray}\label{state_equ1}
 \left\{
    \begin{array}{l}
        \mu_{k}(t)=\mu_k(t-1)-\eta \nabla\mathcal{L}_{\omega_{k}(t)}   + n_{\mu} \\
          \omega_{k}(t)=\mu_{k}(t)+ \tilde{n}_{\mu} 
    \end{array}
\right.
   \end{eqnarray}
where $n_\mu$ is the state noise, and $\tilde{n}_{\mu}$ the observation noise, realizations of $ \mathcal{N}(0, \sigma^2_{\mu})$ and $\mathcal{N}(0, \tilde{\sigma}^2_{\mu})$ respectively. 
   The state and measurement equations for the variance $\sigma_{k}$ are given by:
\begin{eqnarray}\label{state_equ2}
\left\{
\begin{array}{l}
    \sigma_{k}^2(t)=\sigma_{k}^2(t-1)+ \left( \eta \nabla\mathcal{L}_{\omega_{k}(t)}  \right)^2- \eta^2\mu_{k}(t)^2 + n_{\sigma}\\
      z_k(t)=\sigma_{k}^2(t)-\mu_{k}(t)^2+\tilde{n}_{\sigma} \\
      \mbox{ with } z_k(t) =\omega_{k}(t)^2
\end{array}
\right.
   \end{eqnarray}
 \normalsize  
where $n_\sigma$ is the state noise, and $\tilde{n}_{\sigma}$ is the observation noise, realizations of $ \mathcal{N}(0, \sigma^2_{\sigma})$ and $\mathcal{N}(0, \tilde{\sigma}^2_{\sigma})$ respectively. 
As proposed in the main article, and similarly with \cite{ andrychowicz2016learning, yang2019scaling}, we  assume that weights during back-propagation and forward pass are independent. We then get:
\begin{multline}
\Sigma(t)_{k,k'}=\Sigma(t-1)_{k,k'}+\\
\eta^2 \mathbb{E}\left[  \nabla\mathcal{L}_{\omega_{k}(t)}  \nabla\mathcal{L}_{\omega_{k'}(t)}  \right] 
-\eta^2 \mathbb{E}\left[\nabla\mathcal{L}_{\omega_{k}(t)}\right] \mathbb{E}\left[\nabla\mathcal{L}_{\omega_{k'}(t)}\right]
\end{multline}

This leads to the following state and measurement equations for the covariance $\Sigma(t)_{k,k'}$: 
 \begin{eqnarray}\label{state_equ2}
\left\{
\begin{array}{l}
    \Sigma(t)_{k,k'}=\Sigma(t-1)_{k,k'}+ \left( \eta^2 \nabla\mathcal{L}_{\omega_{k}(t)} \nabla\mathcal{L}_{\omega_{k'}(t)} \right))\\- \eta^2\mu_{k}(t)\mu_{k'}(t) + n_{\Sigma}\\
      l_{k,k'}(t)= \Sigma(t)_{k,k'}-\mu_{k}(t)\mu_{k'}(t)+\tilde{n}_{\Sigma} \\
      l_{k,k'}(t)= \Sigma(t)_{k,k'}-\mu_{k}(t)\mu_{k'}(t)+\tilde{n}_{\Sigma} \\
\end{array}
\right.
   \end{eqnarray}
where $n_\Sigma$ is the state noise and $\tilde{n}_{\Sigma}$ is the observation noise, realizations of $ \mathcal{N}(0, \sigma^2_{\Sigma})$ and $\mathcal{N}(0, \tilde{\sigma}^2_{\Sigma})$ respectively. 

\subsection{TRADI training algorithm overview}

We detail the TRADI steps during training in Algorithm \ref{algopseudo}. 

For tracking purposes we must store $\mu_k(t)$ and $\sigma_k(t)$ for all the weights of the network. Hence, we are computationally lighter than Deep Ensembles, which has a training complexity scaling with the number of considered models. In addition, TRADI 
can be applied to any DNN without any modification of the architecture, contrarily to MC dropout that requires adding dropout layers to the underlying DNN.
For clarity we define $ \mathcal{L}(\vomega(t),B(t)) = \frac{1}{|B(t)|} \sum _{ (x_i,y_i) \in B(t)}\ \mathcal{L}(\vomega(t),y_i)$. 
Here $\mathbf{P}_\mu$, $\mathbf{P}_\sigma$ are the noise covariance matrices of the mean and variance respectively and $\mathbf{Q}_\mu$, $\mathbf{Q}_\sigma$ are the optimal gain matrices of the mean and variance respectively. These matrices are used during Kalman filtering \cite{kalman1960new}. 

\begin{figure}
  \caption{TRADI algorithm during training}\label{algopseudo}
  \begin{footnotesize}
\begin{algorithmic}[1]
\State $\vomega(t)$: weights, $\eta$ learning rate, $\sigma_{\mu}$,  $\tilde{\sigma}_{\mu}$, $\sigma_{\sigma}$, $\tilde{\sigma}_{\sigma}$
\State $\mathbf{P}_{\mu}(0)=\mathbf{0}$, $\mathbf{P}_{\sigma}(0)=\mathbf{0}$, $\vomega(0)$, $t=1$
\For{$B(t) \in \mathrm{data}$}
    \State \purple{\textbf{(Forward pass)}} 
    \State $\forall x_i \in B(t)$ calculate $g_{\vomega(t)}(x_i)$ 
    \State evaluate the loss $\mathcal{L}(\vomega(t),B(t)) $ 
    \State \purple{\textbf{(Backward)}}
    \For{$k \in [1:K]$}
        \State  $\omega_{k}(t) \leftarrow \omega_{k}(t-1)-\eta \nabla \mathcal{L}_{\omega_{k}(t)}$
    \EndFor
    \State \purple{\textbf{(Tracking with Kalman filter)}}
    \For{k = 1}{K}
        \State \small{\gray{\emph{\# Update predicted (a priori) estimate covariances}}}
        \State $\mathbf{P}_{\mu}(t^-) \leftarrow \mathbf{P}_{\mu}(t-1)+\sigma_{\mu}$
        \State $\mathbf{P}_{\sigma}(t^-) \leftarrow  \mathbf{P}_{\sigma}(t-1)+\sigma_{\sigma}$
        \State \small{\gray{\emph{\# Update Kalman Gains}}}
        \State $\mathbf{Q}_{\mu} \leftarrow \mathbf{P}_{\mu}(t^-)/(\mathbf{P}_{\mu}(t^-)+\tilde{\sigma}_{\mu})$
        \State $\mathbf{Q}_{\sigma} \leftarrow
        \mathbf{P}_{\sigma}(t^-)/(\mathbf{P}_{\sigma}(t^-)+\tilde{\sigma}_{\sigma})$
        \State \small{\gray{\emph{\# Update mean}}}
        \State  $\mu_{k}(t^-) \leftarrow \mu_{k}(t-1)-\eta \nabla \mathcal{L}_{\omega_{k}(t)}$
        \State  $\mu_{k}(t) \leftarrow (1-\mathbf{Q}_{\mu})\mu_{k}(t^-) + \mathbf{Q}_{\mu} \omega_{k}(t)$
        \State \small{\gray{\emph{\# Update variance}}}
        \State  \small{$\sigma_{k}^2(t^-) \leftarrow \sigma_{k}^2(t-1) +\eta^2\left(\nabla \mathcal{L}_{\omega_{k}(t)}-\mu_{k}(t)\right)^2$}
        \State  $\sigma_{k}^2(t) \leftarrow (1-\mathbf{Q}_{\sigma})\sigma_{k}^2(t^-)+\mathbf{Q}_{\sigma}(\omega_{k}(t)^2 -\mu_{k}(t)^2)$
        \State \small{\gray{\emph{\# Update (a posteriori) estimate covariances}}}
        \State $\mathbf{P}_{\mu}(t) \leftarrow (1-\mathbf{Q}_{\mu})\mathbf{P}_{\mu}(t^-)$
        \State $\mathbf{P}_{\sigma}(t) \leftarrow  (1-\mathbf{Q}_{\sigma})P_{\sigma}(t^-)$
    \EndFor
    \State \purple{\textbf{(Time update)}}
	\State  $t \leftarrow t+1$
\EndFor
  \end{algorithmic}
  \end{footnotesize}
\end{figure}

\subsection{TRADI parameters }

We have set the number of random projections $N=10$ in all experiments in order to get a fast approximation of the covariance matrix. \abc{We should refer again to the paper from Recht or at least make a reference to the main paper on what are these projections.}
We validated this choice experimentally and noticed that the performance is similar for larger values of $N$. $N=10$ ensures a relatively low computational cost. 
We used  $\sigma_{\mbox{\tiny rbf}}=1$ for the RBF parameter of the random projection. We have tested different values, without substantial changes in the results. 
As it can be seen in the algorithm section we have performed a weighted average between the estimated variance/mean with the tracked variance/mean, where the weight depends on Kalman gain.



\section{Complementary results}

In this section, we detail some of the results reported in the main article for the OOD experiments.
The major interest of OOD experiments is that they allow one to see how much we can rely on a DNN. This question is also crucial for industrial research. In this scenario, a particular DNN is trained for a specific application/context which takes into account a certain number of classes. However, in the testing phase new 
unforeseen objects may appear, potentially leading to wrong/dangerous decisions if the DNN confidence is badly calibrated. 



\subsection{Results on MNIST}
\begin{figure}[!th]
     \centering
         \includegraphics[width=0.52\linewidth]{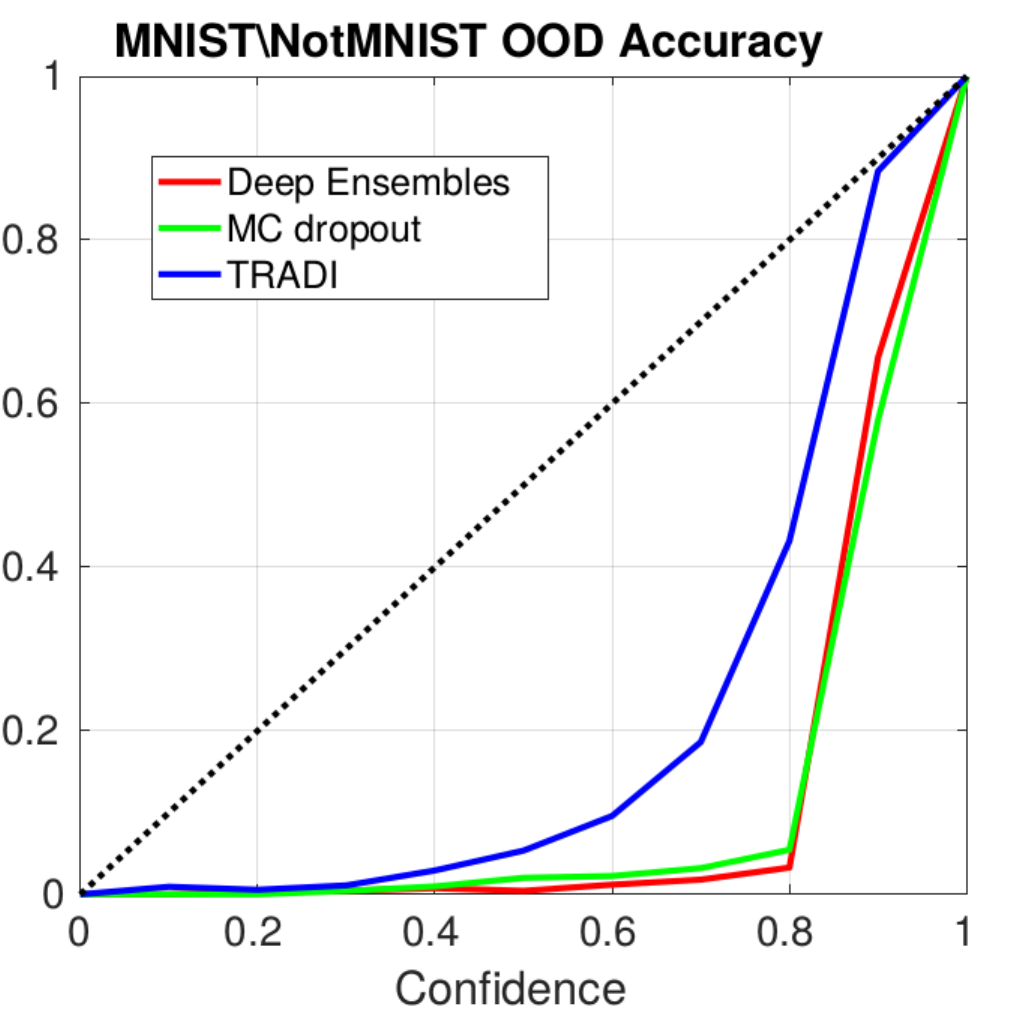}

         \caption{ Calibration plot for MNIST \textbackslash NotMNIST.}
     
\label{mnistacc_cali}
\vspace{-5pt}
\end{figure}

Figure \ref{mnistacc_cali} shows the calibration plots for the OOD experiments with MNIST and ~NotMNIST datasets. As one can see, our strategy (blue curve) has better performances on predicting OOD classes. Calibration plots can easily show whether a DNN is overconfident or not and give an idea on how reliable are the predictions of the DNN. From these plots we see that Deep Ensembles and MC dropout are overconfident, hence they classify non-digits with wrong classes and with high confidence. Our strategy is therefore 
more suitable for this problem, although still improvable in the lower confidence ranges.

\subsection{Results on CamVid}
\begin{figure}[t!]
\renewcommand{\figurename}{Fig.}
\renewcommand{\captionlabelfont}{\bf}
\renewcommand{\captionfont}{\small}
     \centering
        \begin{subfigure}[b]{0.5\linewidth}
        \includegraphics[width=\textwidth]{images/camvid-acc-bins-crop.pdf}		\label{camvid_cali_accu}
        \end{subfigure}\;
        \begin{subfigure}[b]{0.5\linewidth}
        \includegraphics[width=\textwidth]{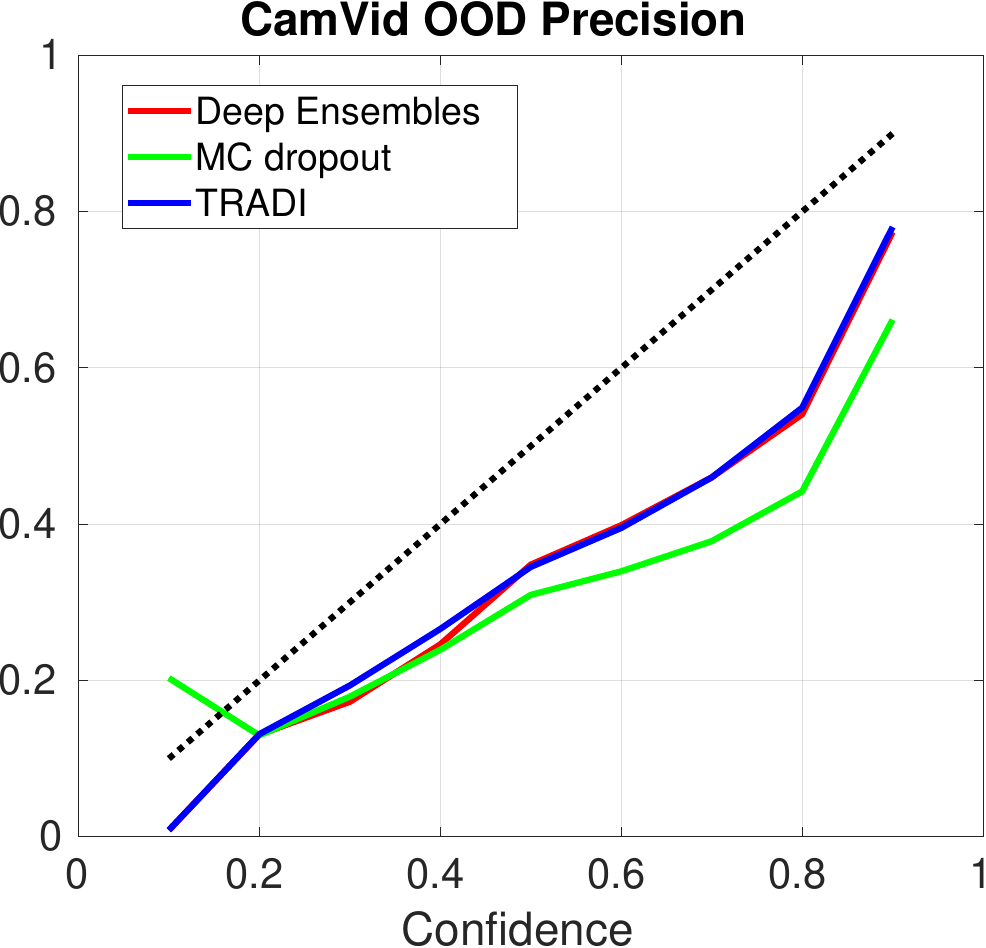}\label{camvid_cali_prec}
        \end{subfigure}\;

         \caption{(a)  Calibration plot for the CamVid experiment.  (b) Calibration plot, where on the Y axis we replace the Accuracy by the average precision of each class for the CamVid experiment.}
\label{fig:all_supp}
\end{figure}

We provide additional scores for CamVid experiments. In Figure \ref{camvid_cali_prec} we illustrate the average precision calibration curve.
This curve is similar to the calibration 
plot, with the different that for each confidence bin, we do not plot the accuracy but the average precision. The usefulness of the precision is that it highlights more the false-positive effects than the accuracy.  
We observe in in Figure \ref{fig:all_supp} that TRADI is better on both measures at identifying OOD classes.

In Table \ref{table:mIoU} we report the mIoU and the global accuracy scores. 
On these metrics, TRADI is between Deep Ensembles and MC Dropout. In contrast to Deep Ensembles we do not need to train multiple DNNs. In order to achieve good performances on semantic segmentation for complex urban scenes, high capacity DNNs are necessary. Training multiple instances of such networks as in Deep Ensembles brings a significant computational cost. Furthermore, these models are usually updated when new data is recorded and each time the full ensemble needs updating. TRADI requires training a single network each time.  

\begin{table}[t!]
\centering
\begin{tabular}{|l|l|l|l|}
\hline
              & MC dropout & Deep Ensembles & TRADI  \\ \hline
mean IoU      & 0.4857   &\textbf{ 0.5719  }      & 0.5298 \\ \hline
Accuracy   & 0.8034   &\textbf{ 0.8806 }       & 0.8488 \\ \hline
\end{tabular}
\caption{CamVid semantic segmentation results (mIoU, accuracy). }\label{table:mIoU}
\vspace{-7pt}
\end{table}

\begin{figure}[!thb]
     \centering
         \includegraphics[width=0.52\linewidth]{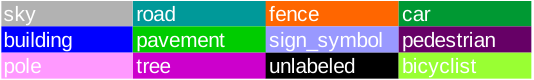}
         \caption{Color map for the CamVid experiment.}
     \label{CamVid_colormap}
\vspace{-5pt}
\end{figure}
\begin{figure}[htb]
\begin{center}
\begin{tabular}{l l l l }
\includegraphics[width=0.25\columnwidth]{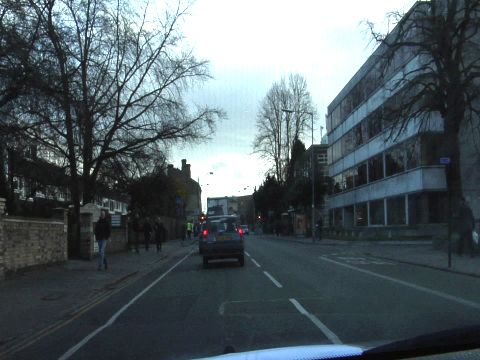}&
\includegraphics[width=0.25\columnwidth]{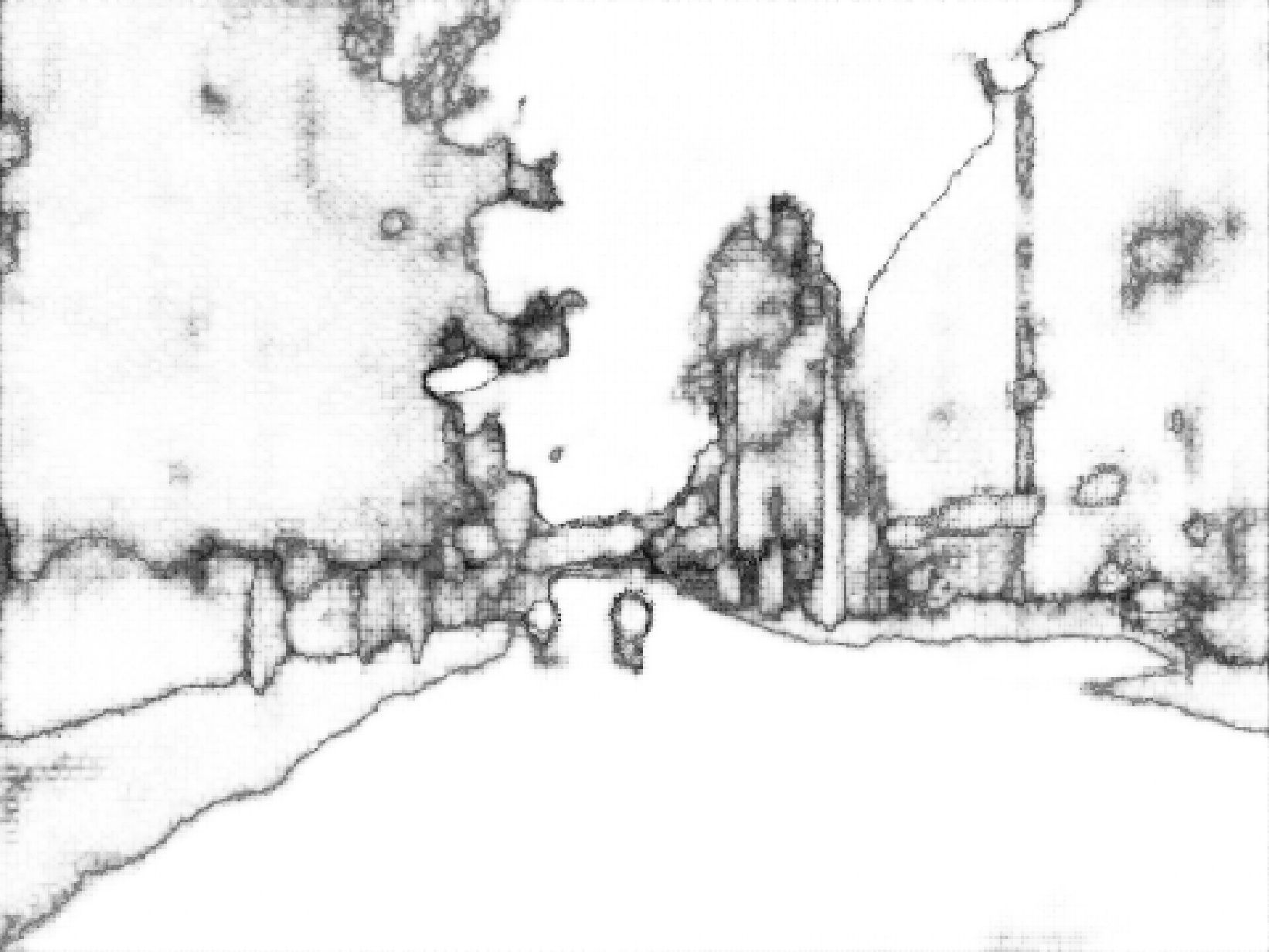}&
\includegraphics[width=0.25\columnwidth]{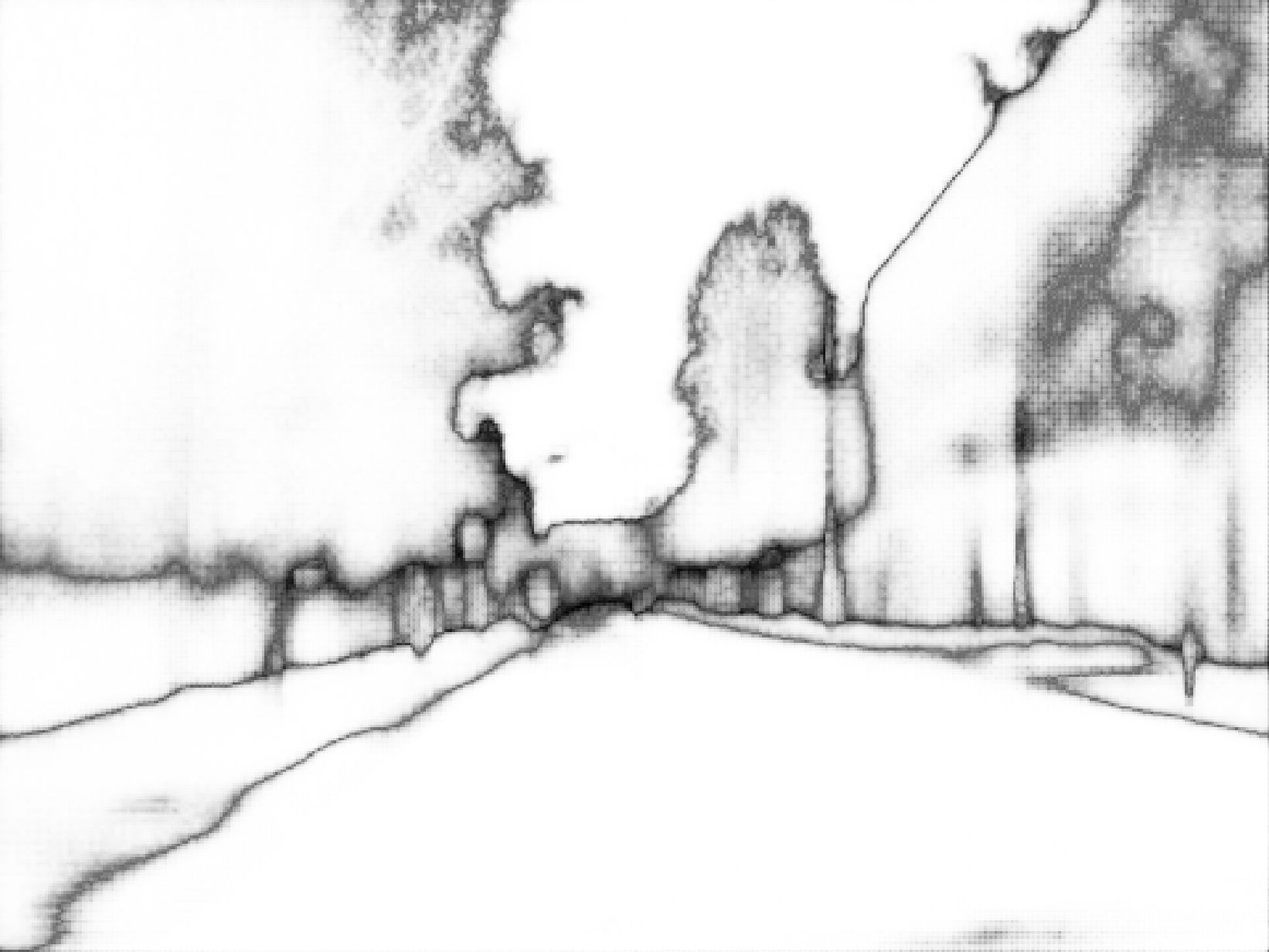}&
\includegraphics[width=0.25\columnwidth]{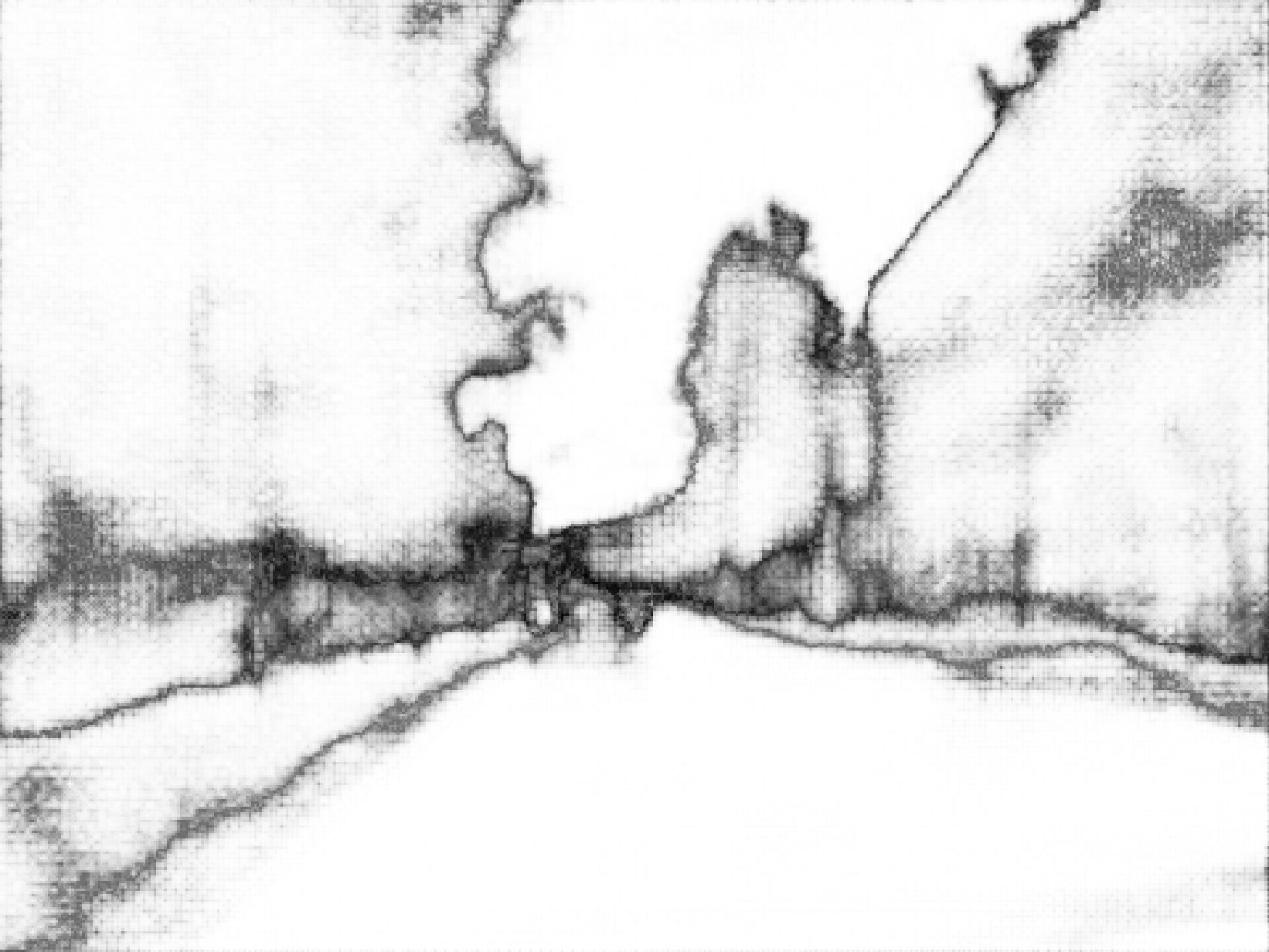}\\
\includegraphics[width=0.25\columnwidth]{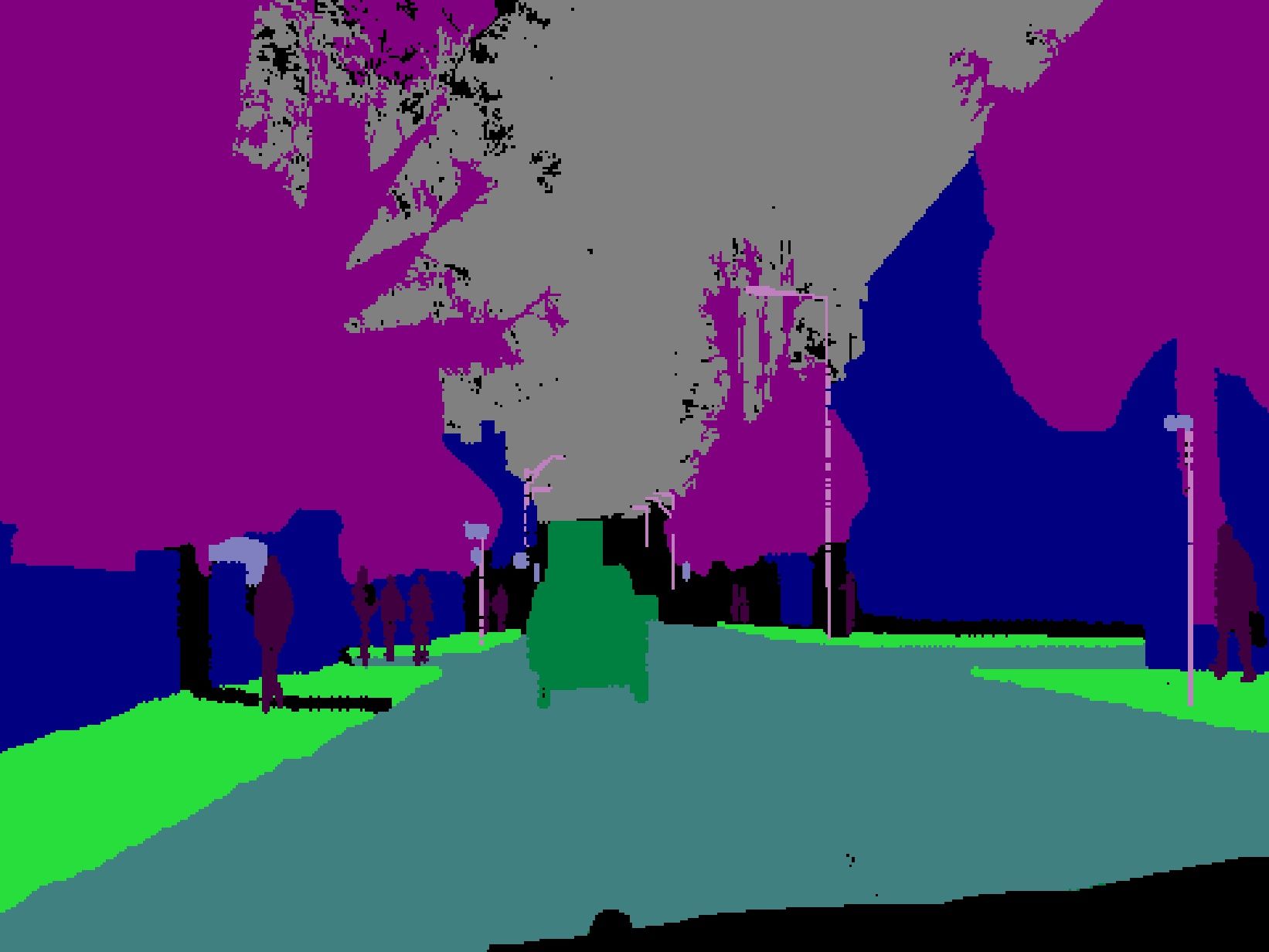} &
\includegraphics[width=0.25\columnwidth]{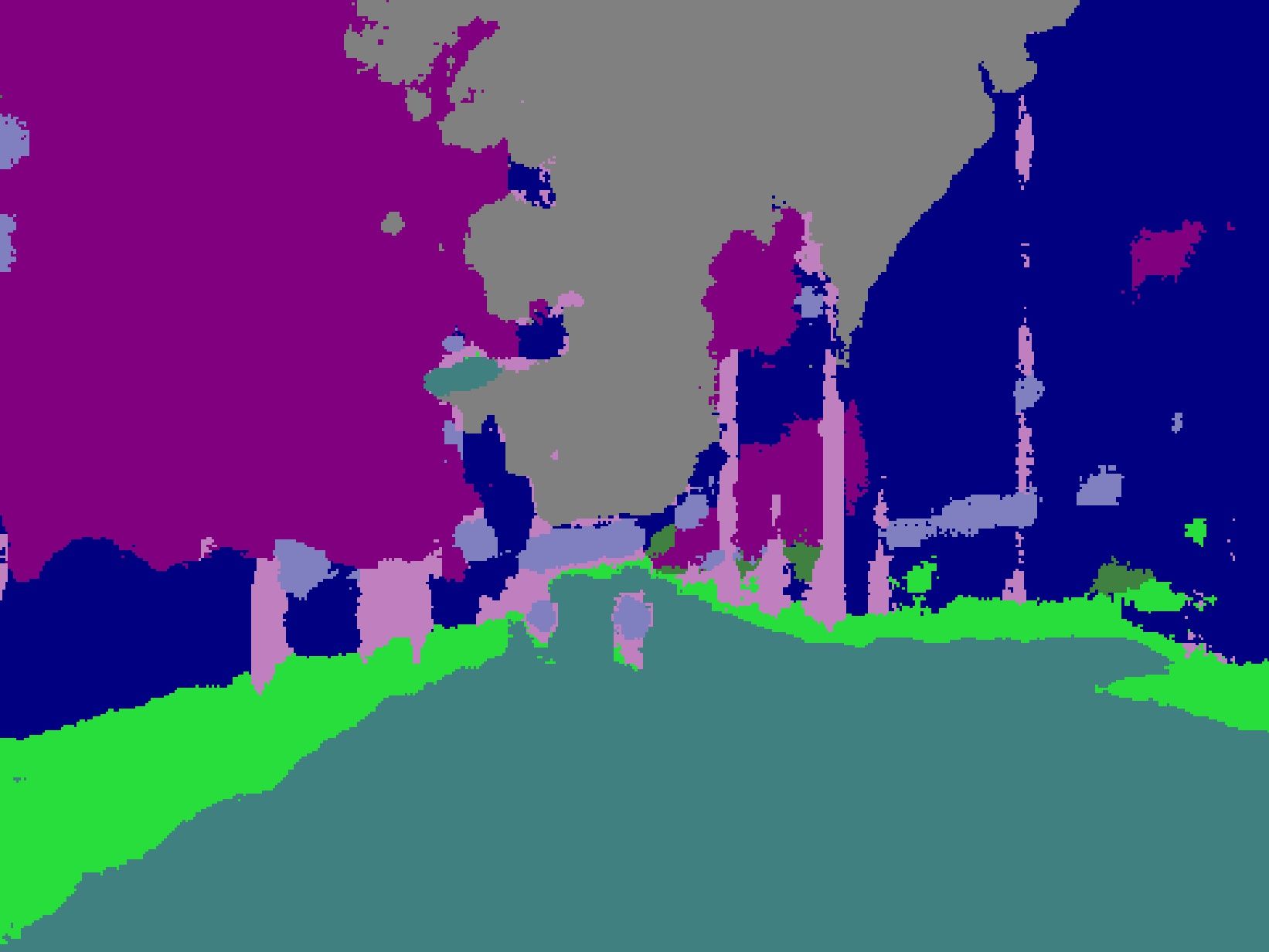} &
\includegraphics[width=0.25\columnwidth]{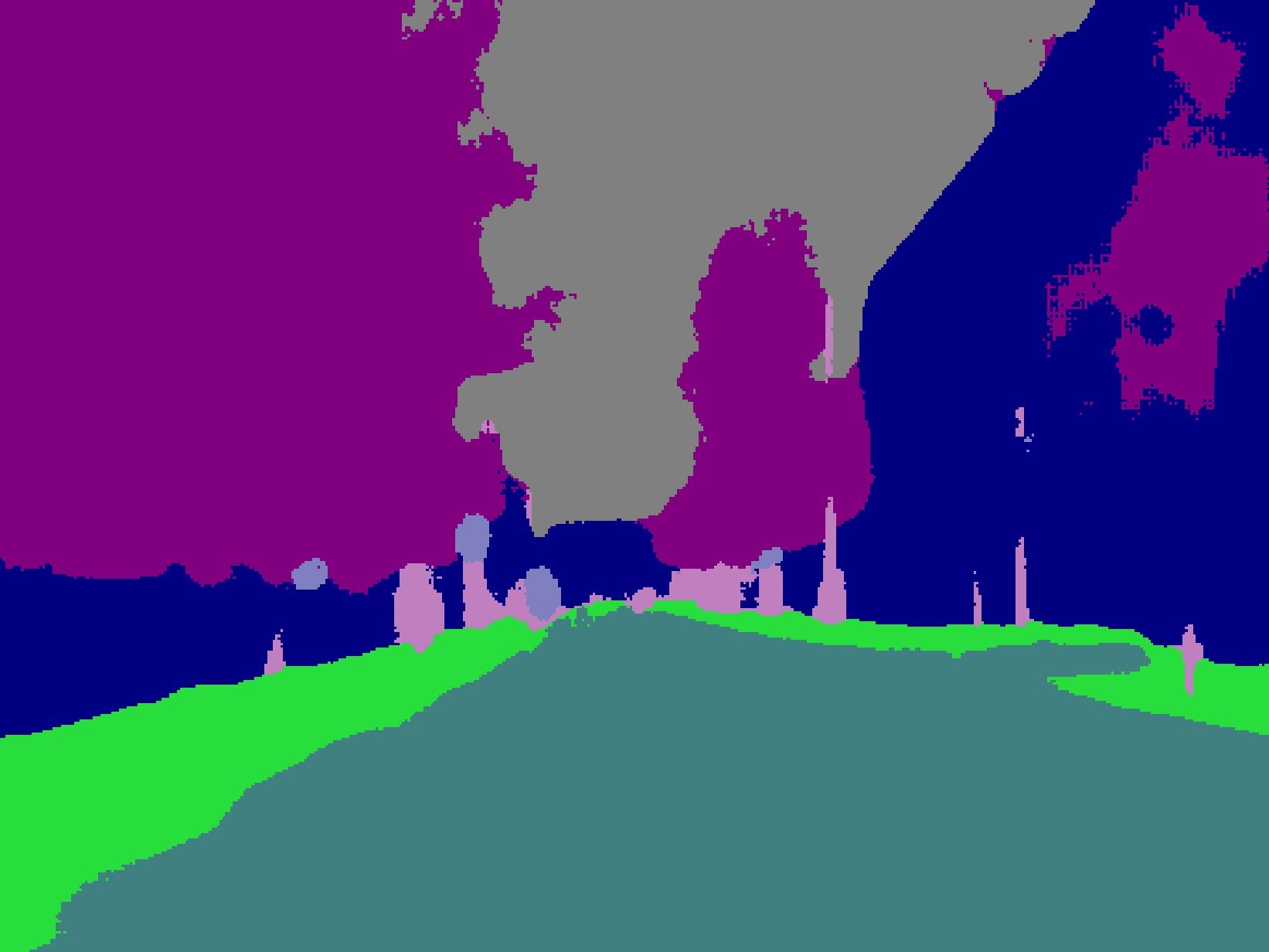}&
\includegraphics[width=0.25\columnwidth]{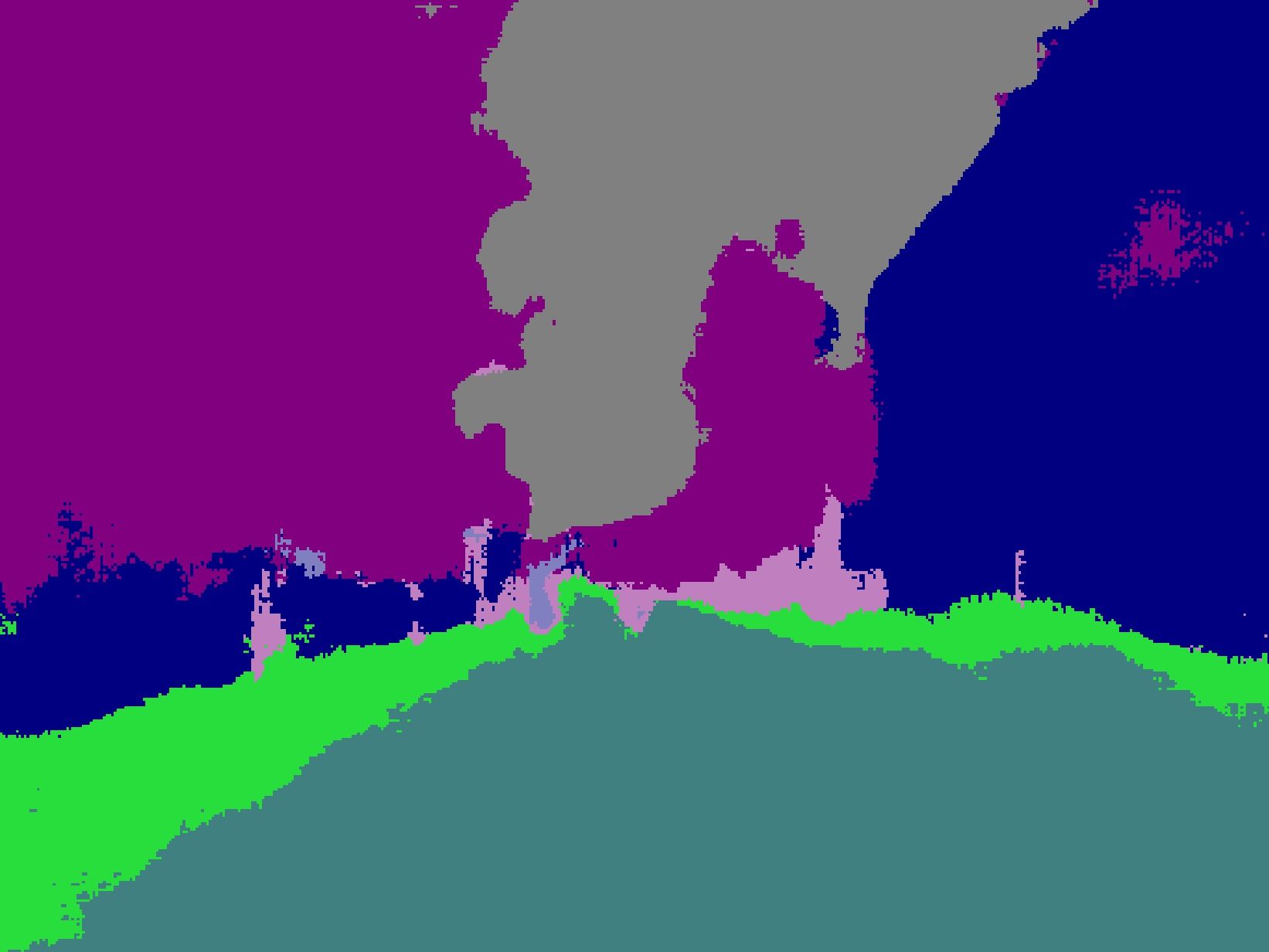}\\
\end{tabular}
\end{center}
\caption{ Qualitative results on CamVid experiments. Column (1): \emph{top} - input image (the image contrast has been enhanced for clarity with respect to the original dataset image), \emph{bottom} - ground truth; Columns (2-4): \emph{top} - confidence  scores from MC dropout, Deep Ensembles and TRADI respectively, \emph{bottom} - corresponding segmentation predictions.}
\label{fig:accuconfidanc_img2_landscape}
\end{figure}
\begin{figure}[htb]
\begin{center}
\begin{tabular}{l l l l}
\includegraphics[width=0.25\columnwidth]{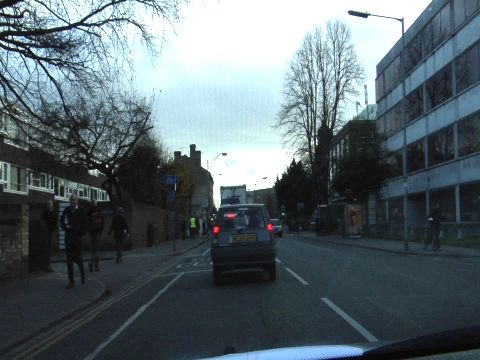}&
\includegraphics[width=0.25\columnwidth]{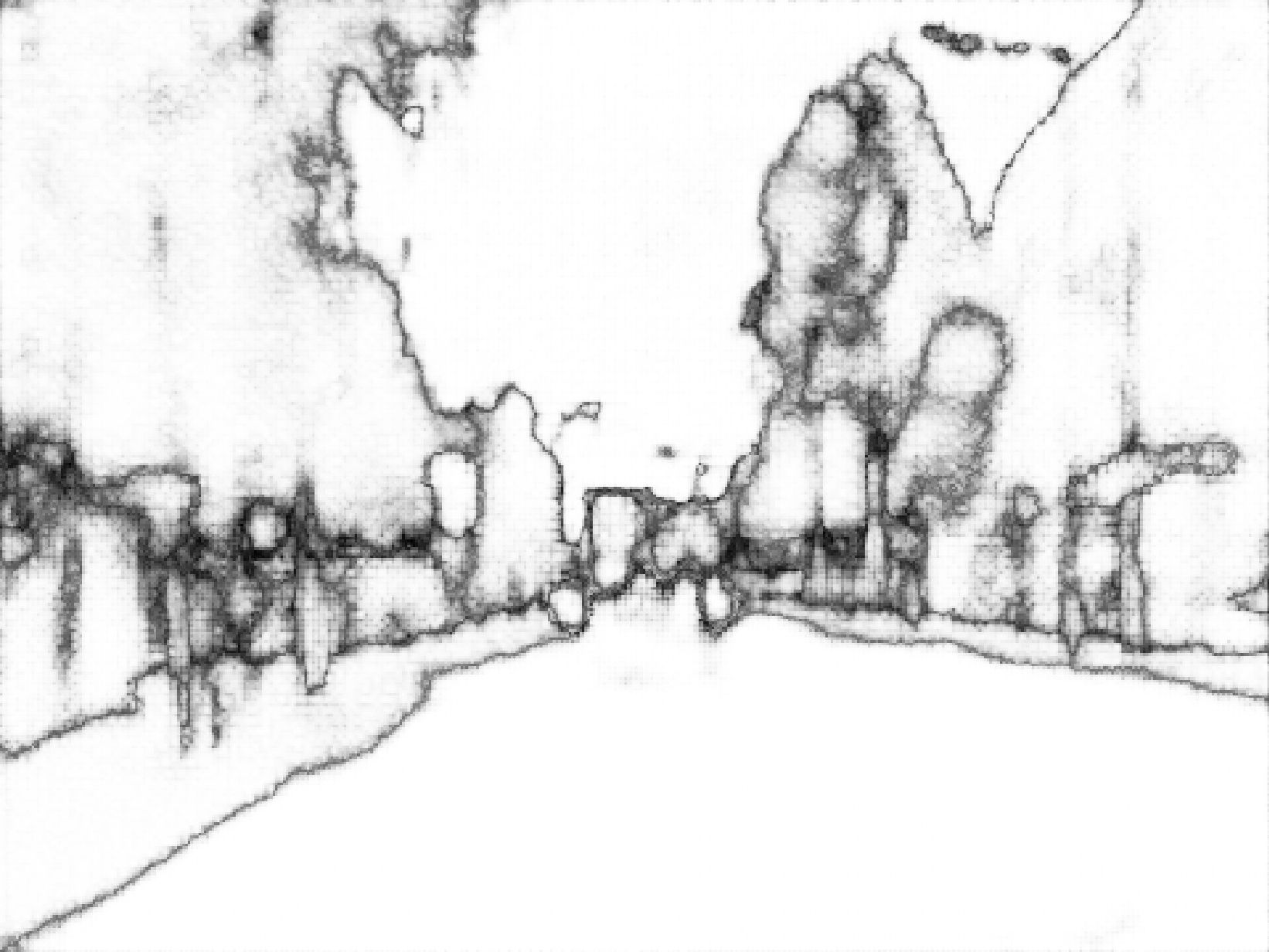}&
\includegraphics[width=0.25\columnwidth]{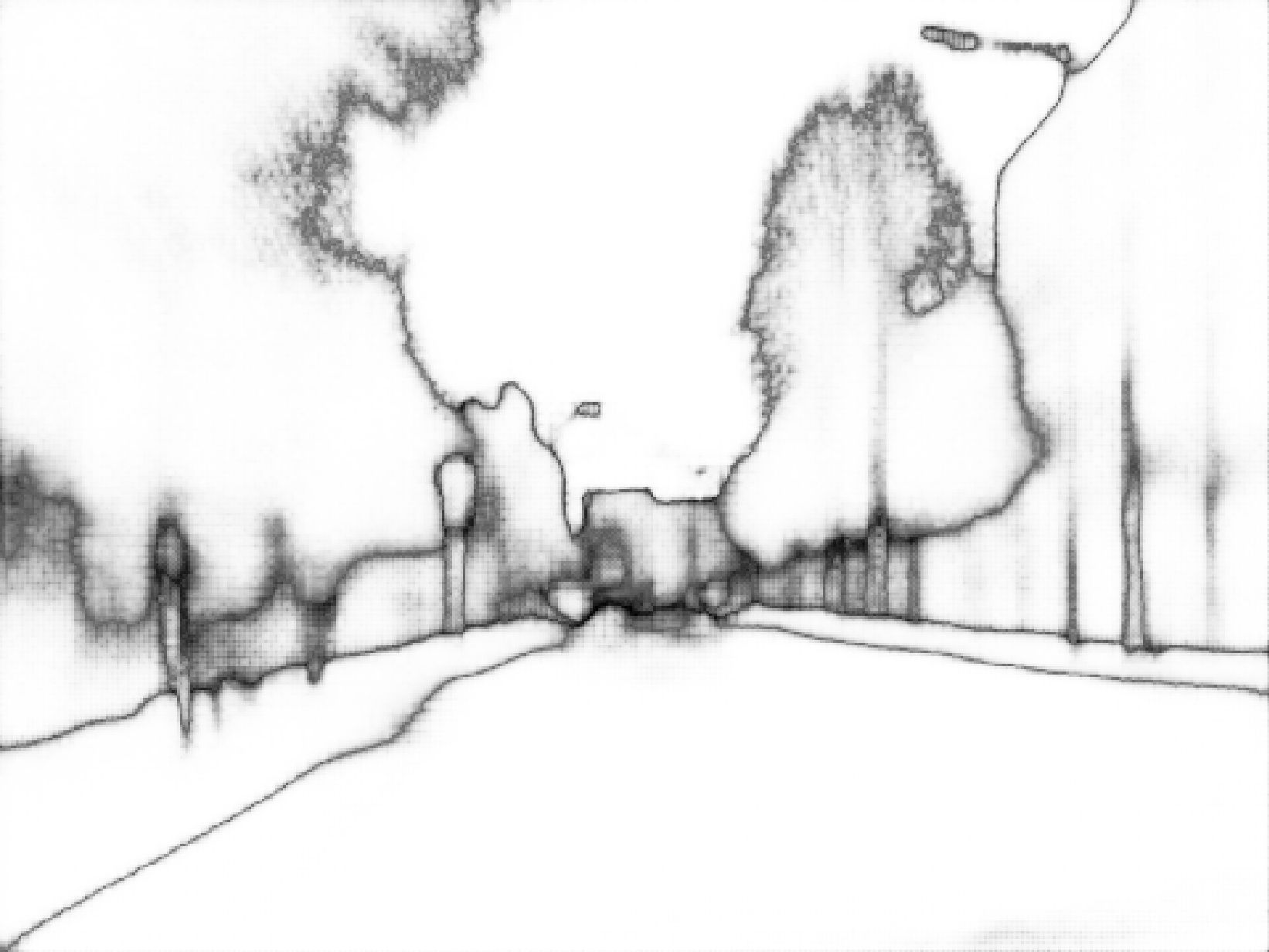}&
\includegraphics[width=0.25\columnwidth]{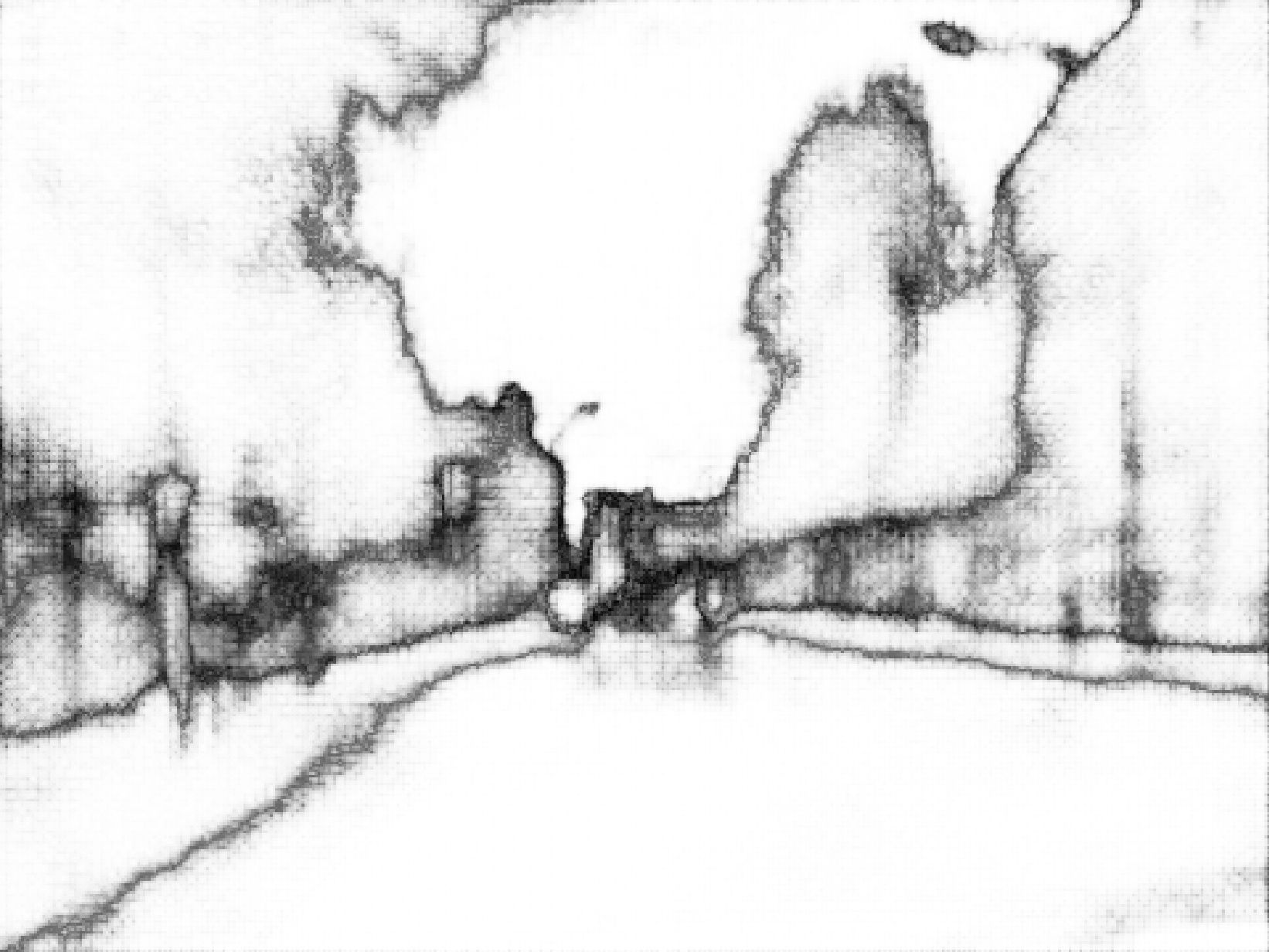}\\
\includegraphics[width=0.25\columnwidth]{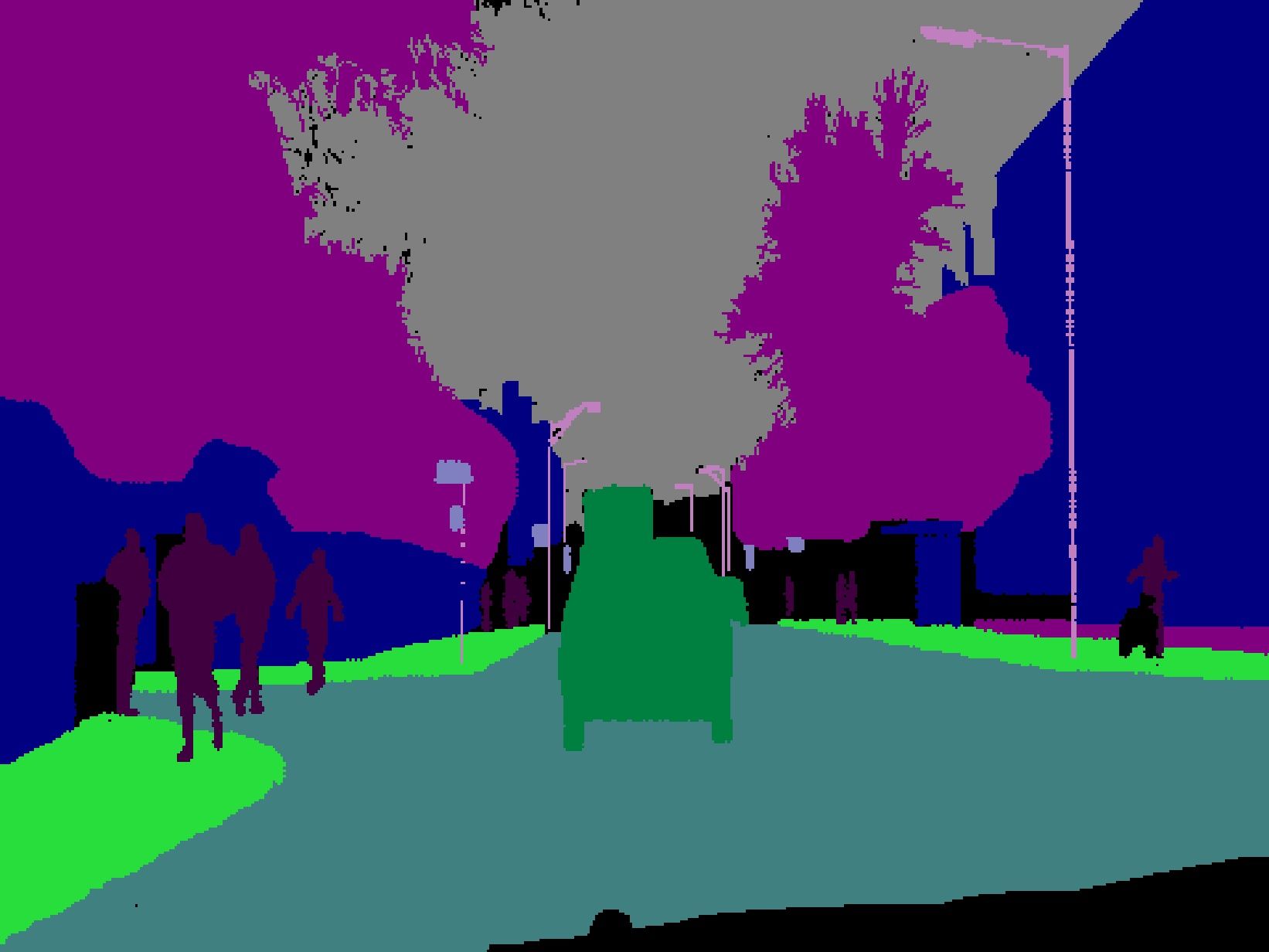}&
\includegraphics[width=0.25\columnwidth]{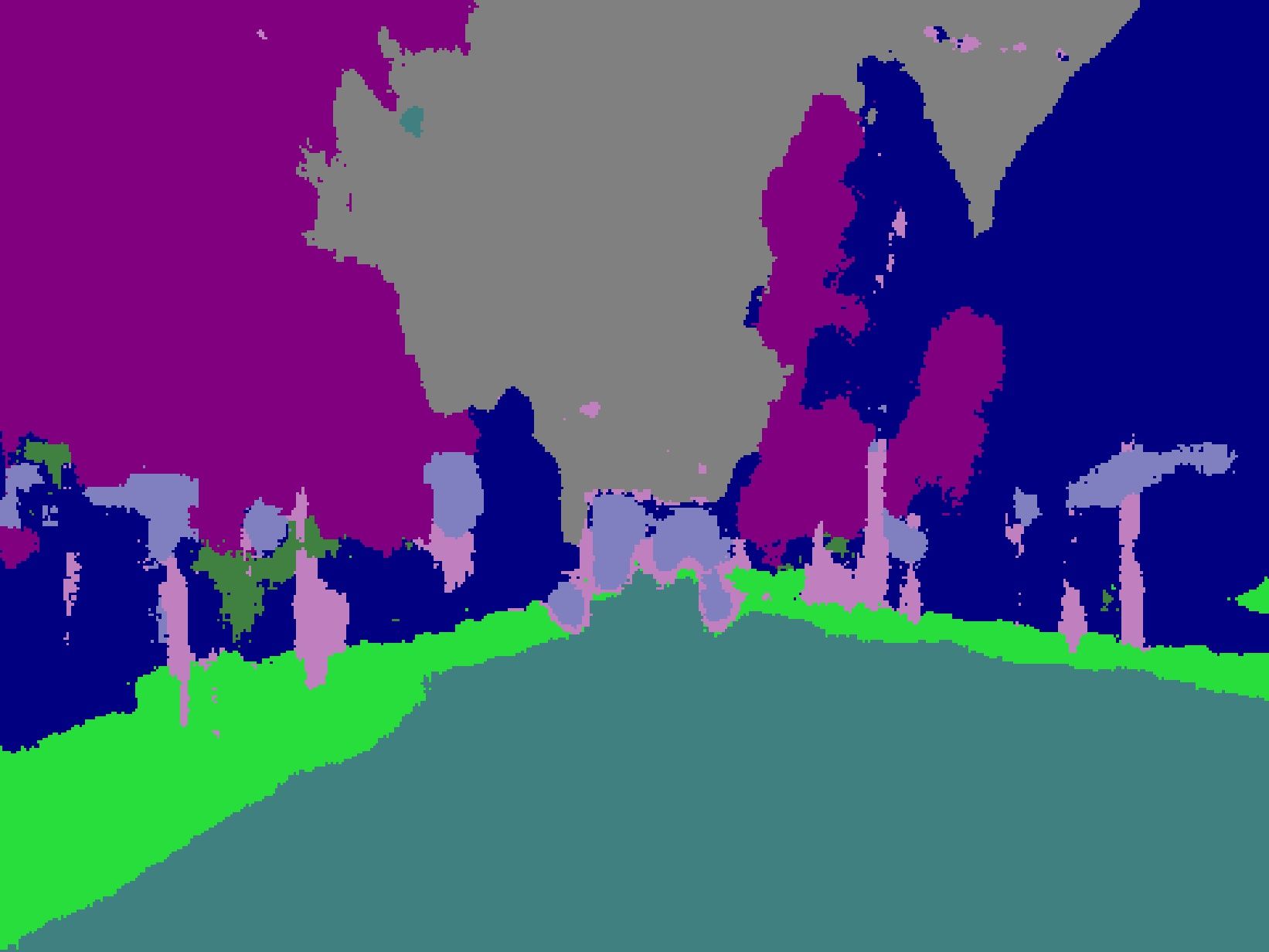}&
\includegraphics[width=0.25\columnwidth]{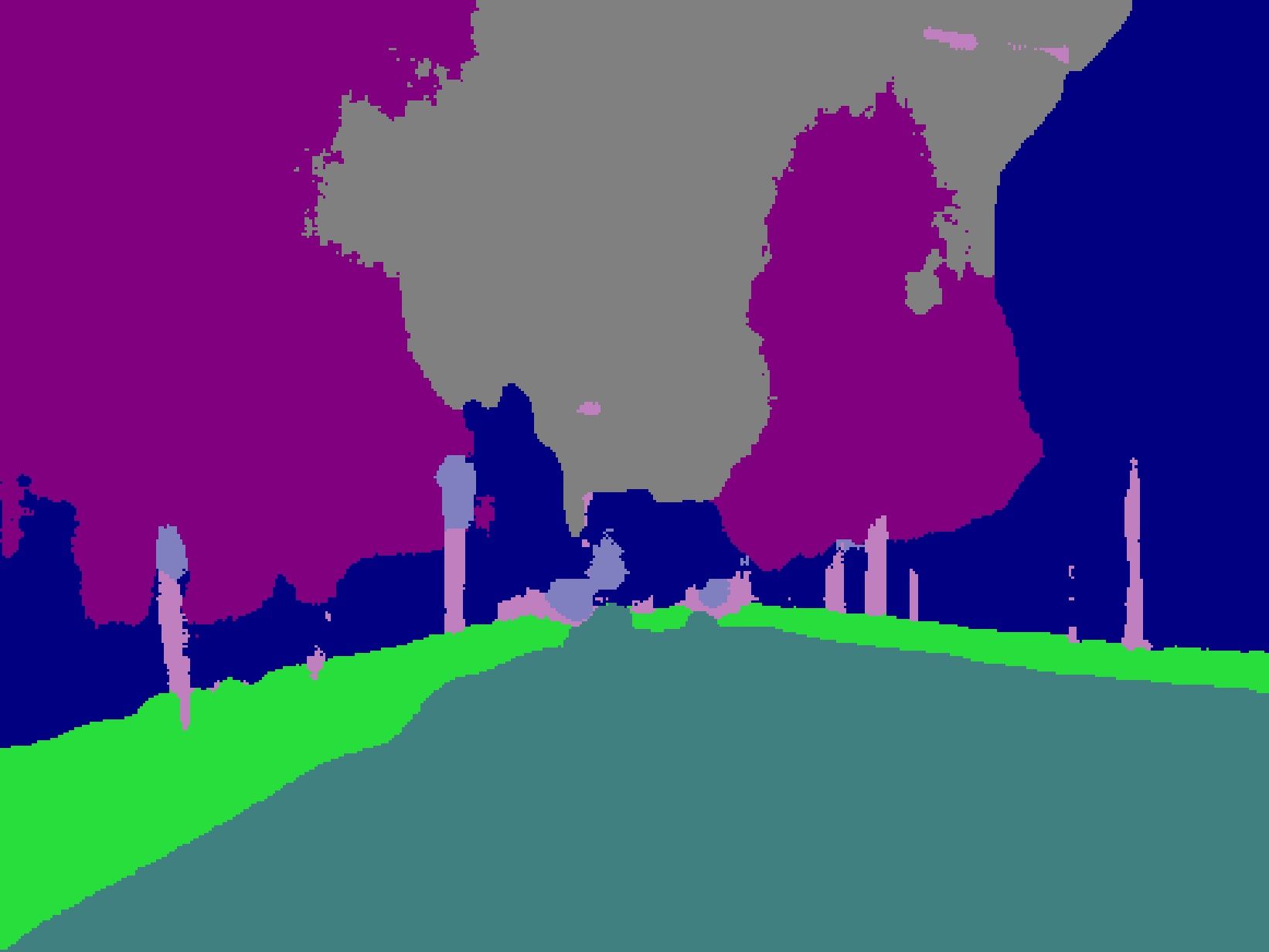}&
\includegraphics[width=0.25\columnwidth]{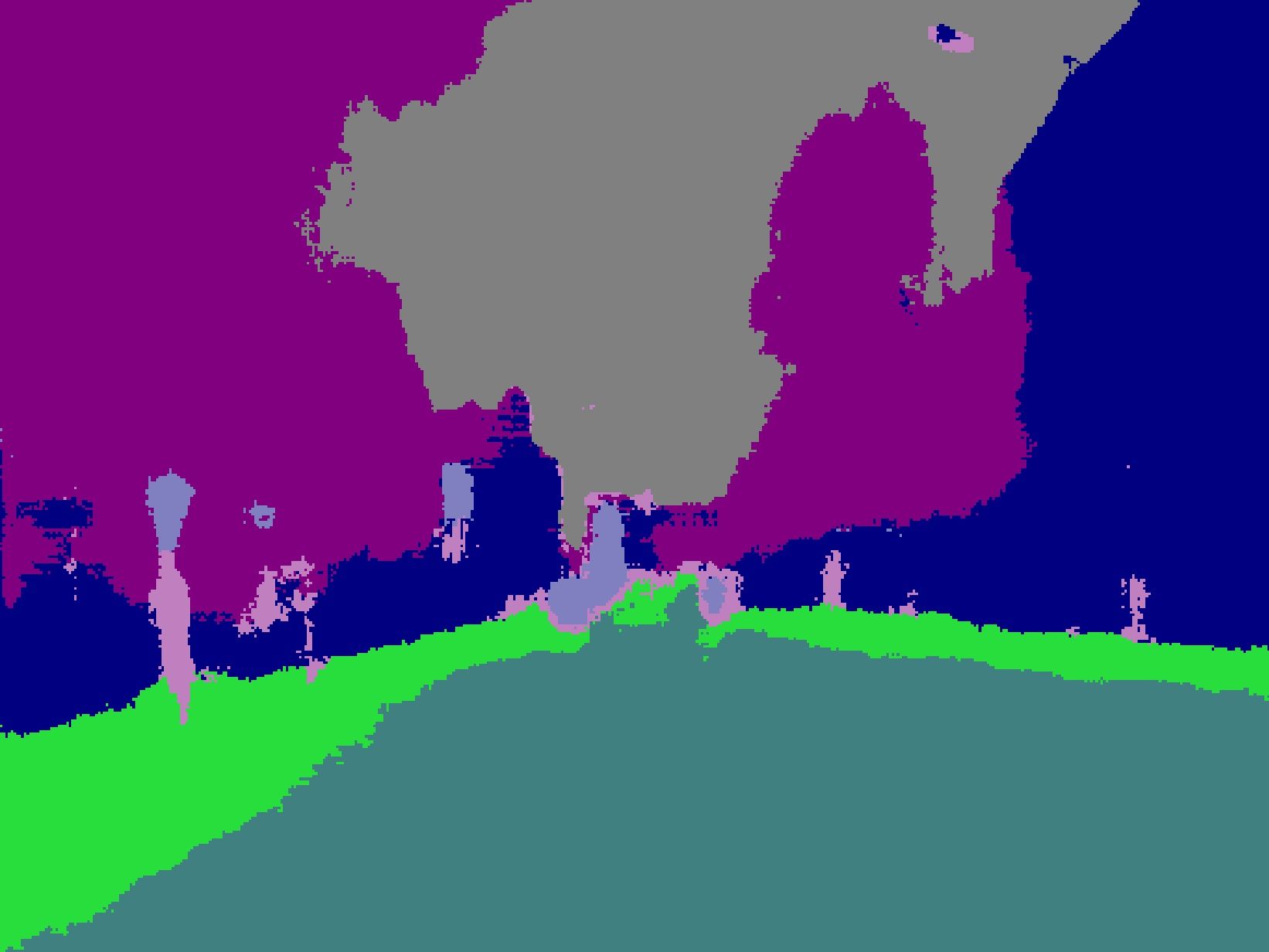}\\
\end{tabular}
\end{center}
\caption{Qualitative results on CamVid experiments. Column (1): \emph{top} - input image (the image contrast has been enhanced for clarity with respect to the original dataset image), \emph{bottom} - ground truth; Columns (2-4): \emph{top} - confidence  scores from MC dropout, Deep Ensembles and TRADI respectively, \emph{bottom} - corresponding segmentation predictions.}
 \label{fig:accuconfidanc_img3_landscape}
\end{figure}

In Figures~\ref{fig:accuconfidanc_img2_landscape}, \ref{fig:accuconfidanc_img3_landscape}, \ref{fig:accuconfidanc_img2_zoom_landscape}, and \ref{fig:accuconfidanc_img3_zoom_landscape} we report additional qualitative results. Figures~\ref{fig:accuconfidanc_img2_zoom_landscape} and~\ref{fig:accuconfidanc_img3_zoom_landscape} show zoom-in over areas of interest in Figures~\ref{fig:accuconfidanc_img2_landscape} and \ref{fig:accuconfidanc_img3_zoom_landscape} respectively.
We provide the color code for the semantic segmentation map in Figure~\ref{CamVid_colormap}. We remind that in this experiments the classes human, bicyclist, and car are used as OOD and removed from the train set. 
We can see that TRADI outputs less confident predictions for human pixels, comparing to Deep Ensebles and MC Dropout.

\vspace{-5pt}
\paragraph{Comparing with Deep Ensembles.} 
Deep Ensembles is among the most powerful and effective techniques for epistemic uncertainty. However few works on uncertainty estimation with DNNs on computer vision tasks have considered it for evaluation. We argue that this work is one of the first to truly challenge Deep Ensembles. 
While we do not achieve higher accuracy than Deep Ensembles, our approach strikes a good compromise between computational cost for training and prediction performance. The computational budget for Deep Ensembles is proportional to the number of models in the ensemble, while for TRADI we always train a single model regardless of the number of network samples we have at test time. Our results on the OOD experiments challenge and sometimes outperform the Deep Ensembles ones. 


\begin{figure}[htb]
\begin{center}
\begin{tabular}{l l l l }
\includegraphics[trim={1.7cm 3.2cm 9.8cm 5.8cm},clip,width=0.25\columnwidth]{./images/enhanced_4_img2.jpg}&
\includegraphics[trim={2cm 4cm 12cm 7cm},clip,width=0.25\columnwidth]{./images/4_confidence_drop2.jpg}&
\includegraphics[trim={2cm 4cm 12cm 7cm},clip,width=0.25\columnwidth]{./images/4_confidence_deep_ense2.jpg}&
\includegraphics[trim={2cm 4cm 12cm 7cm},clip,width=0.25\columnwidth]{./images/4_confidence_tradi2.jpg}\\
\includegraphics[trim={2cm 4cm 12cm 7cm},clip,width=0.25\columnwidth]{./images/4_labels.jpg}&
\includegraphics[trim={2cm 4cm 12cm 7cm},clip,width=0.25\columnwidth]{./images/4_prediction_drop2.jpg}&
\includegraphics[trim={2cm 4cm 12cm 7cm},clip,width=0.25\columnwidth]{./images/4_prediction_deep_ense2.jpg}&
\includegraphics[trim={2cm 4cm 12cm 7cm},clip,width=0.25\columnwidth]{./images/4_prediction_tradi2.jpg}\\
\end{tabular}
\end{center}
\caption{ Detailed qualitative analysis of the confidence on CamVid predictions. Column (1): top - input image, bottom - ground truth; Columns (2-4): top - confidence  scores from MC dropout, Deep Ensembles and TRADI respectively, bottom - corresponding segmentation predictions.}
 \label{fig:accuconfidanc_img2_zoom_landscape}
   \end{figure}

\begin{figure}[htb]
\begin{center}
\begin{tabular}{l l l l }
\includegraphics[trim={1.7cm 3.2cm 9.8cm 5.8cm},clip,width=0.25\columnwidth]{./images/enhanced_6_img2.jpg}&
\includegraphics[trim={2cm 4cm 12cm 7cm},clip,width=0.25\columnwidth]{./images/6_confidence_drop2.jpg}&
\includegraphics[trim={2cm 4cm 12cm 7cm},clip,width=0.25\columnwidth]{./images/6_confidence_deep_ense2.jpg}&
\includegraphics[trim={2cm 4cm 12cm 7cm},clip,width=0.25\columnwidth]{./images/6_confidence_tradi2.jpg}\\
\includegraphics[trim={2cm 4cm 12cm 7cm},clip,width=0.25\columnwidth]{./images/6_labels.jpg}&
\includegraphics[trim={2cm 4cm 12cm 7cm},clip,width=0.25\columnwidth]{./images/6_prediction_drop2.jpg}&
\includegraphics[trim={2cm 4cm 12cm 7cm},clip,width=0.25\columnwidth]{./images/6_prediction_deep_ense2.jpg}&
\includegraphics[trim={2cm 4cm 12cm 7cm},clip,width=0.25\columnwidth]{./images/6_prediction_tradi2.jpg}\\
\end{tabular}
\end{center}
\caption{ Detailed qualitative analysis of the confidence on CamVid predictions. Column (1): top - input image, bottom - ground truth; Columns (2-4): top - confidence  scores from MC dropout, Deep Ensembles and TRADI respectively, bottom - corresponding segmentation predictions.}
 \label{fig:accuconfidanc_img3_zoom_landscape}
\end{figure}

%% file: main.bbl
\begin{thebibliography}{10}
\providecommand{\url}[1]{\texttt{#1}}
\providecommand{\urlprefix}{URL }
\providecommand{\doi}[1]{https://doi.org/#1}

\bibitem{NotMnist}
Notmnist dataset.
  \url{http://yaroslavvb.blogspot.com/2011/09/notmnist-dataset.html}

\bibitem{andrychowicz2016learning}
Andrychowicz, M., Denil, M., Gomez, S., Hoffman, M.W., Pfau, D., Schaul, T.,
  Shillingford, B., De~Freitas, N.: Learning to learn by gradient descent by
  gradient descent. In: Advances in neural information processing systems. pp.
  3981--3989 (2016)

\bibitem{beluch2018power}
Beluch, W.H., Genewein, T., N{\"u}rnberger, A., K{\"o}hler, J.M.: The power of
  ensembles for active learning in image classification. In: Proceedings of the
  IEEE Conference on Computer Vision and Pattern Recognition. pp. 9368--9377
  (2018)

\bibitem{bishop2006pattern}
Bishop, C.M.: Pattern recognition and machine learning. springer (2006)

\bibitem{blundell15bnn}
Blundell, C., Cornebise, J., Kavukcuoglu, K., Wierstra, D.: Weight uncertainty
  in neural network. In: Bach, F., Blei, D. (eds.) Proceedings of the 32nd
  International Conference on Machine Learning. Proceedings of Machine Learning
  Research, vol.~37, pp. 1613--1622. PMLR, Lille, France (07--09 Jul 2015),
  \url{http://proceedings.mlr.press/v37/blundell15.html}

\bibitem{blundell2015weight}
Blundell, C., Cornebise, J., Kavukcuoglu, K., Wierstra, D.: Weight uncertainty
  in neural networks. arXiv preprint arXiv:1505.05424  (2015)

\bibitem{brostow2008segmentation}
Brostow, G.J., Shotton, J., Fauqueur, J., Cipolla, R.: Segmentation and
  recognition using structure from motion point clouds. In: European conference
  on computer vision. pp. 44--57. Springer (2008)

\bibitem{ionet2018}
Chen, C., Lu, C.X., Markham, A., Trigoni, N.: Ionet: Learning to cure the curse
  of drift in inertial odometry. In: The Thirty-Second AAAI Conference on
  Artificial Intelligence (AAAI-18) (2018)

\bibitem{dosovitskiy17}
Dosovitskiy, A., Ros, G., Codevilla, F., Lopez, A., Koltun, V.: {CARLA}: {An}
  open urban driving simulator. In: Proceedings of the 1st Annual Conference on
  Robot Learning. pp. 1--16 (2017)

\bibitem{gal2016dropout}
Gal, Y., Ghahramani, Z.: Dropout as a bayesian approximation: Representing
  model uncertainty in deep learning. In: international conference on machine
  learning. pp. 1050--1059 (2016)

\bibitem{gal2017concrete}
Gal, Y., Hron, J., Kendall, A.: Concrete dropout. In: NIPS (2017)

\bibitem{glorot2010understanding}
Glorot, X., Bengio, Y.: Understanding the difficulty of training deep
  feedforward neural networks. In: Proceedings of the thirteenth international
  conference on artificial intelligence and statistics. pp. 249--256 (2010)

\bibitem{graves2011practical}
Graves, A.: Practical variational inference for neural networks. In: Advances
  in neural information processing systems. pp. 2348--2356 (2011)

\bibitem{grewal2011kalman}
Grewal, M.S.: Kalman filtering. Springer (2011)

\bibitem{guo2017calibration}
Guo, C., Pleiss, G., Sun, Y., Weinberger, K.Q.: On calibration of modern neural
  networks. In: Proceedings of the 34th International Conference on Machine
  Learning-Volume 70. pp. 1321--1330. JMLR. org (2017)

\bibitem{haarnoja2016backprop}
Haarnoja, T., Ajay, A., Levine, S., Abbeel, P.: Backprop kf: Learning
  discriminative deterministic state estimators. In: Advances in Neural
  Information Processing Systems. pp. 4376--4384 (2016)

\bibitem{he2015delving}
He, K., Zhang, X., Ren, S., Sun, J.: Delving deep into rectifiers: Surpassing
  human-level performance on imagenet classification. In: Proceedings of the
  IEEE international conference on computer vision. pp. 1026--1034 (2015)

\bibitem{he2016deep}
He, K., Zhang, X., Ren, S., Sun, J.: Deep residual learning for image
  recognition. In: Proceedings of the IEEE conference on computer vision and
  pattern recognition. pp. 770--778 (2016)

\bibitem{hendrycks2019anomalyseg}
Hendrycks, D., Basart, S., Mazeika, M., Mostajabi, M., Steinhardt, J., Song,
  D.: A benchmark for anomaly segmentation. arXiv preprint arXiv:1911.11132
  (2019)

\bibitem{hendrycks2016baseline}
Hendrycks, D., Gimpel, K.: A baseline for detecting misclassified and
  out-of-distribution examples in neural networks. arXiv preprint
  arXiv:1610.02136  (2016)

\bibitem{hernandez2015probabilistic}
Hern{\'a}ndez-Lobato, J.M., Adams, R.: Probabilistic backpropagation for
  scalable learning of bayesian neural networks. In: International Conference
  on Machine Learning. pp. 1861--1869 (2015)

\bibitem{ioffe2015batch}
Ioffe, S., Szegedy, C.: Batch normalization: Accelerating deep network training
  by reducing internal covariate shift. arXiv preprint arXiv:1502.03167  (2015)

\bibitem{izmailov2018averaging}
Izmailov, P., Podoprikhin, D., Garipov, T., Vetrov, D., Wilson, A.G.: Averaging
  weights leads to wider optima and better generalization. arXiv preprint
  arXiv:1803.05407  (2018)

\bibitem{kalman1960new}
Kalman, R.E.: A new approach to linear filtering and prediction problems.
  Journal of basic Engineering  \textbf{82}(1),  35--45 (1960)

\bibitem{kendall2015bayesian}
Kendall, A., Badrinarayanan, V., , Cipolla, R.: Bayesian segnet: Model
  uncertainty in deep convolutional encoder-decoder architectures for scene
  understanding. arXiv preprint arXiv:1511.02680  (2015)

\bibitem{kendall2017uncertainties}
Kendall, A., Gal, Y.: What uncertainties do we need in bayesian deep learning
  for computer vision? In: Advances in neural information processing systems.
  pp. 5574--5584 (2017)

\bibitem{kingma2014adam}
Kingma, D.P., Ba, J.: Adam: A method for stochastic optimization. arXiv
  preprint arXiv:1412.6980  (2014)

\bibitem{kingma13vae}
Kingma, D.P., Welling, M.: Auto-encoding variational bayes. In: 2nd
  International Conference on Learning Representations, {ICLR} 2014, Banff, AB,
  Canada, April 14-16, 2014, Conference Track Proceedings (2014)

\bibitem{krizhevsky2009learning}
Krizhevsky, A., Hinton, G., et~al.: Learning multiple layers of features from
  tiny images. Tech. rep., Citeseer (2009)

\bibitem{krizhevsky2012imagenet}
Krizhevsky, A., Sutskever, I., Hinton, G.E.: Imagenet classification with deep
  convolutional neural networks. In: Advances in neural information processing
  systems. pp. 1097--1105 (2012)

\bibitem{lakshminarayanan2017simple}
Lakshminarayanan, B., Pritzel, A., Blundell, C.: Simple and scalable predictive
  uncertainty estimation using deep ensembles. In: Advances in Neural
  Information Processing Systems. pp. 6402--6413 (2017)

\bibitem{lambert2018}
Lambert, J., Sener, O., Savarese, S.: Deep learning under privileged
  information using heteroscedastic dropout. 2018 IEEE/CVF Conference on
  Computer Vision and Pattern Recognition pp. 8886--8895 (2018)

\bibitem{lan2019lca}
Lan, J., Liu, R., Zhou, H., Yosinski, J.: Lca: Loss change allocation for
  neural network training. In: Advances in Neural Information Processing
  Systems. pp. 3614--3624 (2019)

\bibitem{lecun1998gradient}
LeCun, Y., Bottou, L., Bengio, Y., Haffner, P., et~al.: Gradient-based learning
  applied to document recognition. Proceedings of the IEEE  \textbf{86}(11),
  2278--2324 (1998)

\bibitem{lee2015m}
Lee, S., Purushwalkam, S., Cogswell, M., Crandall, D., Batra, D.: Why m heads
  are better than one: Training a diverse ensemble of deep networks. arXiv
  preprint arXiv:1511.06314  (2015)

\bibitem{liu2019neural}
Liu, C., Gu, J., Kim, K., Narasimhan, S.G., Kautz, J.: Neural rgb (r) d
  sensing: Depth and uncertainty from a video camera. In: Proceedings of the
  IEEE Conference on Computer Vision and Pattern Recognition. pp. 10986--10995
  (2019)

\bibitem{maddox2019simple}
Maddox, W., Garipov, T., Izmailov, P., Vetrov, D., Wilson, A.G.: A simple
  baseline for bayesian uncertainty in deep learning. arXiv preprint
  arXiv:1902.02476  (2019)

\bibitem{mukhoti2018eval}
Mukhoti, J., Gal, Y.: Evaluating bayesian deep learning methods for semantic
  segmentation. CoRR  \textbf{abs/1811.12709} (2018),
  \url{http://arxiv.org/abs/1811.12709}

\bibitem{neal1996}
Neal, R.M.: Bayesian Learning for Neural Networks. Springer-Verlag, Berlin,
  Heidelberg (1996)

\bibitem{yann}
Ollivier, Y.: The extended kalman filter is a natural gradient descent in
  trajectory space. arXiv preprint arXiv:1901.00696  (2019)

\bibitem{osband2016}
Osband, I.: Risk versus uncertainty in deep learning : Bayes , bootstrap and
  the dangers of dropout (2016)

\bibitem{osband2018}
Osband, I., Aslanides, J., Cassirer, A.: Randomized prior functions for deep
  reinforcement learning. In: NeurIPS (2018)

\bibitem{paszke2016enet}
Paszke, A., Chaurasia, A., Kim, S., Culurciello, E.: Enet: A deep neural
  network architecture for real-time semantic segmentation. arXiv preprint
  arXiv:1606.02147  (2016)

\bibitem{paszke2019pytorch}
Paszke, A., Gross, S., Massa, F., Lerer, A., Bradbury, J., Chanan, G., Killeen,
  T., Lin, Z., Gimelshein, N., Antiga, L., et~al.: Pytorch: An imperative
  style, high-performance deep learning library. In: Advances in Neural
  Information Processing Systems. pp. 8024--8035 (2019)

\bibitem{Rahimi2007}
Rahimi, A., Recht, B.: Random features for large-scale kernel machines. In:
  Advances in neural information processing systems. pp. 1177--1184 (2007)

\bibitem{salimans2016weight}
Salimans, T., Kingma, D.P.: Weight normalization: A simple reparameterization
  to accelerate training of deep neural networks. In: Advances in Neural
  Information Processing Systems. pp. 901--909 (2016)

\bibitem{simonyan2014very}
Simonyan, K., Zisserman, A.: Very deep convolutional networks for large-scale
  image recognition. arXiv preprint arXiv:1409.1556  (2014)

\bibitem{srivastava2014dropout}
Srivastava, N., Hinton, G., Krizhevsky, A., Sutskever, I., Salakhutdinov, R.:
  Dropout: A simple way to prevent neural networks from overfitting. J. Mach.
  Learn. Res.  \textbf{15}(1),  1929--1958 (Jan 2014),
  \url{http://dl.acm.org/citation.cfm?id=2627435.2670313}

\bibitem{szegedy2014going}
Szegedy, C., Liu, W., Jia, Y., Sermanet, P., Reed, S., Anguelov, D., Erhan, D.,
  Vanhoucke, V., Rabinovich, A., et~al.: Going deeper with convolutions. arxiv
  2014. arXiv preprint arXiv:1409.4842  \textbf{1409} (2014)

\bibitem{teye2018}
Teye, M., Azizpour, H., Smith, K.: Bayesian uncertainty estimation for batch
  normalized deep networks. In: ICML (2018)

\bibitem{wang2018batch}
Wang, G., Peng, J., Luo, P., Wang, X., Lin, L.: Batch kalman normalization:
  Towards training deep neural networks with micro-batches. arXiv preprint
  arXiv:1802.03133  (2018)

\bibitem{williams2006gaussian}
Williams, C.K., Rasmussen, C.E.: Gaussian processes for machine learning,
  vol.~2. MIT press Cambridge, MA (2006)

\bibitem{yang2019scaling}
Yang, G.: Scaling limits of wide neural networks with weight sharing: Gaussian
  process behavior, gradient independence, and neural tangent kernel
  derivation. arXiv preprint arXiv:1902.04760  (2019)

\bibitem{yu2018bdd100k}
Yu, F., Xian, W., Chen, Y., Liu, F., Liao, M., Madhavan, V., Darrell, T.:
  Bdd100k: A diverse driving video database with scalable annotation tooling.
  arXiv preprint arXiv:1805.04687  (2018)

\bibitem{zagoruyko2016wide}
Zagoruyko, S., Komodakis, N.: Wide residual networks. arXiv preprint
  arXiv:1605.07146  (2016)

\bibitem{zhao2017pyramid}
Zhao, H., Shi, J., Qi, X., Wang, X., Jia, J.: Pyramid scene parsing network.
  In: Proceedings of the IEEE conference on computer vision and pattern
  recognition. pp. 2881--2890 (2017)

\end{thebibliography}
